\documentclass[a4paper,fleqn]{cas-dc}

\usepackage[utf8]{inputenc}
\usepackage{tabu}
\usepackage{times}
\usepackage[T1]{fontenc}
\usepackage{graphicx}
\graphicspath{{images}}
\usepackage{multirow}
\usepackage{adjustbox}
\usepackage{hyperref}

\usepackage{caption}
\usepackage{subcaption}
\usepackage{algorithmic}

\usepackage{url}
\usepackage{amsmath,amssymb,amsfonts}
\usepackage{textcomp}
\usepackage{color,soul}
\usepackage{multirow}

\usepackage[numbers]{natbib}

\def\tsc#1{\csdef{#1}{\textsc{\lowercase{#1}}\xspace}}
\tsc{WGM}
\tsc{QE}
\tsc{EP}
\tsc{PMS}
\tsc{BEC}
\tsc{DE}

\begin{document}



\title [mode = title]{P2AT: Pyramid Pooling Axial Transformer for Real-time Semantic Segmentation}                      

\tnotetext[1]{This work is supported by the National Natural Science Foundation of China (Nos.62272418,62102058), Basic public welfare research program of Zhejiang Province(No.LGG18E050011).}


\author{Mohammed A. M. Elhassan\textsuperscript{a,c},Changjun Zhou\textsuperscript{a,*},Amina Benabid\textsuperscript{b,c}, Abuzar B. M. Adam\textsuperscript{d}}

\address[1]{School of Computer Science, Zhejiang Normal University,Jinhua,321004,Zhejiang,P.R.China}
\address[2]{School of Mathematical Medicine, Zhejiang Normal University,Jinhua,321004,Zhejiang,P.R.China}
\address[3]{Zhejiang Institute of Photoelectronics \& Zhejiang Institute for Advanced Light Source, Zhejiang Normal University}
\address[4]{School of Communications and Information Engineering, Chongqing University of Posts and Telecommunications, Chongqing,40065,P.R.China}

\cortext[cor1]{Corresponding authors\\
				mohammedac29@zjnu.edu.cn (Mohammed A. M. Elhassan)\\
				zhouchangjun@zjnu.edu.cn (Changjun Zhou)\\
				amina.benabid@zjnu.edu.cn (Amina Benabid)\\
				abuzar@cqupt.edu.cn(Abuzar B. M. Adam)}

\begin{abstract}
Recently, Transformer-based models have achieved promising results in various vision tasks, due to their ability to model long-range dependencies. However, transformers are computationally expensive, which limits their applications in real-time tasks such as autonomous driving. In addition, an efficient local and global feature selection and fusion are vital for accurate dense prediction, especially driving scene understanding tasks. In this paper, we propose a real-time semantic segmentation architecture named Pyramid Pooling Axial Transformer (P2AT). The proposed P2AT takes a coarse feature from the CNN encoder to produce scale-aware contextual features, which are then combined with the multi-level feature aggregation scheme to produce enhanced contextual features. Specifically, we introduce a pyramid pooling axial transformer to capture intricate spatial and channel dependencies, leading to improved performance on semantic segmentation. Then, we design a Bidirectional Fusion module (BiF) to combine semantic information at different levels. Meanwhile, a Global Context Enhancer is introduced to compensate for the inadequacy of concatenating different semantic levels. Finally, a decoder block is proposed to help maintain a larger receptive field. We evaluate P2AT variants on three challenging scene-understanding datasets. In particular, our P2AT variants achieve state-of-art results on the Camvid dataset 80.5\%, 81.0\%, 81.1\% for P2AT-S, P2AT-M, and P2AT-L, respectively. Furthermore, our experiment on Cityscapes and Pascal VOC 2012 have demonstrated the efficiency of the proposed architecture,  with results showing that P2AT-M, achieves 78.7\% on Cityscapes. The source code will be available at \footnote{\url{https://github.com/mohamedac29/P2AT}}.
\end{abstract}




\maketitle

\section{Introduction}
\label{sec:introduction}
Perception is a vital task of any intelligent driving system, which collects the necessary information about the surrounding environment of the moving vehicle. As an inseparable part of automatic driving, visual perception is being explored and researched by major mainstream automobile manufacturers, enterprises, universities, and scientific research institutes. The large-scale application of artificial intelligence in the automotive industry accelerates the development in this field. 
High precision and speed architectures are crucial for the future development of advanced driver assistance systems and autonomous vehicles. The Research in visual perception algorithms based on deep learning is a very important part of landing industrial technology applications because deep learning methods have unique abilities for constructing robust intelligent driving algorithms in many research directions, such as traffic sign recognition \cite{min2022traffic,tian2019traffic}, lane detection \cite{Huang2023,Wang2023}, target detection \cite{carion2020end,han2019small}, driving free space recognition and semantic segmentation \cite{chen2017deeplab,paszke2016enet}. Fast and accurate semantics segmentation and target detection are prerequisites for safe, intelligent driving. 

\begin{figure}
    \centering
	\includegraphics[width=1\linewidth]{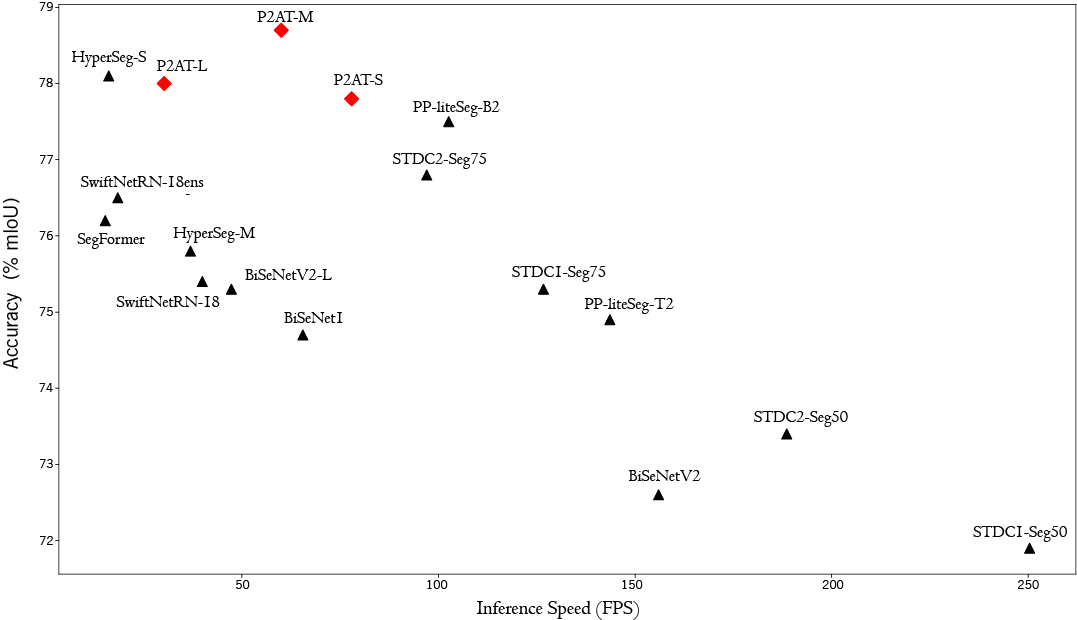}
	\caption{The inference speed and accuracy trade-off for real-time models on the Cityscapes \cite{cordts2016cityscapes} test set. 
          red color refers to our models, while black represents others.}
	\label{Fig:refinement_module}
\end{figure} 

As the complexity of image content increases, the task of semantic segmentation becomes progressively more challenging due to intricate structures and variations in color, texture, and scale. In recent years, deep learning has significantly influenced the development of various semantic segmentation approaches, emerging as the dominant framework. Many state-of-the-art methods for semantic segmentation, such as those described in \cite{chen2017deeplab,zhao2017pyramid,bilinski2018dense,zhao2018psanet}, utilize Fully Convolutional Networks (FCNs) \cite{long2015fully} as fundamental components. Notably, PSPNet \cite{zhao2017pyramid} and Deeplab \cite{chen2017deeplab} introduce specialized modules for capturing multi-scale contextual information, namely the Pyramid Pooling Module (PPM) and the Atrous Spatial Pyramid Module (ASPP), respectively. Despite these advancements, challenges persist, particularly when handling complex image content, as existing approaches tend to generate imprecise masks. 

In recent years, the remarkable performance of Vision Transformer (ViT) in image classification \cite{dosovitskiy2020image,touvron2021training} has spurred efforts to extend its application to semantic segmentation tasks. These endeavors have significantly improved over previous semantic segmentation convolutional neural networks (CNNs) \cite{liu2021swin,liu2022swin,xie2021segformer,zheng2021rethinking}. However, implementing pure transformer models for semantic segmentation has a considerable computational cost, particularly when dealing with large input images. To address this issue, \cite{liu2021swin} has introduced hierarchical Vision Transformers, which offer a more computationally efficient alternative. SegFormer \cite{xie2021segformer} has proposed refined design for the encoder and decoder, resulting in an efficient semantic segmentation ViT. Nevertheless, one concern with SegFormer is its heavy reliance on increasing the model capacity of the encoder as the primary means of improving performance, potentially limiting its overall efficiency. 

Unlike the aforementioned methods that introduced pure Transformer for dense pixel prediction, we propose a hybrid architecture for better and more efficient semantic segmentation of autonomous driving. Specifically, since contextual information is crucial for semantic segmentation, we exploit the Pyramid Pooling Axial Transformer in CNN to capture global context information effectively. To fully utilize the merit of Transformer and CNN, a bidirectional fusion module is proposed to integrate the feature from the network encoder and global context information, which is then refined using the global context enhancer.  As a hybrid ConvNet-Transformer framework, our P2AT can accurately segment objects in the scene for autonomous driving with faster inference speed.


Our main contributions are summarized as follows:\\
\begin{enumerate}
    \item We introduce a novel pyramid pooling Axial Transformer framework (P2AT) for real-time semantic segmentation. To achieve an accuracy/speed trade-off, four modules, including, Scale-aware context aggregator module, multi-level feature fusion module, a decoder , and feature refinement module are designed, leading to the following contributions. 
	\item We encapsulate pyramid pooling to the Axial Transformer to extract contextual features, leading to powerful architecture that is easier to train in small datasets.  
	\item We introduce a multi-level fusion module to fuse encoded detailed representations and deep semantic features. Specifically, a bidirectional fusion (BiF) module based on semantic feature upsampler (SFU) and local feature refinement (LFR) is designed to obtain efficient feature fusion.
 \item We introduce a global context enhancer (GCE) module to compensate for the inadequacy of concatenating different semantic levels.
	\item We propose an efficient decoder based on enhanced ConvNext and a feature refinement module proposed to remove noises to enhance the final prediction.
	\item We evaluate P2AT on three challenging scene understanding datasets: Camvid, Cityscapes, and PASCAL VOC 2012. The results show that P2AT achieves state-of-the-art results.
\end{enumerate}

\begin{figure*}
	\centering
	\includegraphics[width=0.95\linewidth]{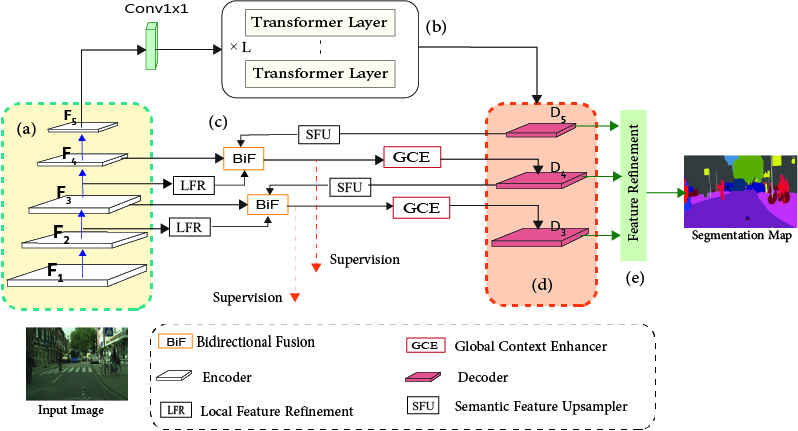}
	\caption{The architecture of P2AT. (a) Encoder based on
pre-trained ResNet, (b) Transformer Layers to extract contextual information, (c) Multi-stage Feature Fusion Block (d) Decoder Block (e) Feature Refinement Block.}
	\label{p2at}
\end{figure*}

\section{Related Work}
\subsection{Semantic Segmentation}

Semantic segmentation has witnessed significant advancements in the deep learning era \cite{krizhevsky2017imagenet,he2016deep,simonyan2014very}. The introduction of Fully Convolutional Networks (FCNs) \cite{long2015fully} revolutionized semantic segmentation by enabling end-to-end pixel-to-pixel classification. Building upon FCNs, researchers have explored various avenues to enhance semantic segmentation performance.  Efforts have been dedicated to enlarging the receptive field \cite{zhao2017pyramid,chen2014semantic,yu2015multi,he2016deep,chen2017deeplab,yang2018denseaspp}, incorporating boundary information \cite{ding2019boundary,bertasius2016semantic,li2020improving}, refining contextual information \cite{yuan2018ocnet,yu2020context,yuan2020object,lin2017refinenet,9292449cgnet}, integrating attention modules \cite{elhassan2021dsanet, fu2019dual,huang2019ccnet,zhao2018psanet,li2018pyramid}, and leveraging AutoML technologies \cite{shaw2019squeezenas,chen2019fasterseg,liu2019auto} to design optimized models for scene parsing. These approaches have substantially improved the accuracy of semantic segmentation. More recently, Transformer-based architectures have demonstrated their efficacy in semantic segmentation \cite{zheng2021rethinking,yan2022lawin}. Nevertheless, these methods still require significant computational resources, which limits their applications in real-time tasks such as autonomous driving.
To speed up the segmentation and reduce the computational cost, ICNet \cite{zhao2018icnet} proposes a cascade network with multi-resolution input image. DFANet \cite{li2019dfanet} utilizes a lightweight backbone to speed up its network and proposes a cross-level feature aggregation to boost accuracy. SwiftNet \cite{orsic2019defense} uses lateral connections as the cost-effective solution to restore the prediction resolution while maintaining the speed. BiSeNet \cite{yu2021bisenet}, ContextNet \cite{poudel2018contextnet}, GUN \cite{mazzini2018guided}, and DSANet \cite{elhassan2021dsanet} introduce spatial and semantic paths to reduce computation. SFNet \cite{li2020semantic} aligns feature maps from adjacent levels and further enhances the feature maps using a feature pyramid framework. ESPNet \cite{mehta2018espnet} save computation by decomposing standard convolution into point-wise convolution and spatial pyramid of dilated convolutions. 

\textbf{Vision Transformer:} Originally introduced in Natural Language Processing (NLP) tasks, have gained significant traction in computer vision. These models, relying on self-attention mechanisms, excel at capturing long-range dependencies among tokens in sentences. Moreover, transformers possess inherent parallelization capabilities, enabling efficient training on extensive datasets. Inspired by the success of transformers in NLP, several methodologies have emerged in computer vision that combines Convolutional Neural Networks (CNNs) with various forms of self-attention to tackle diverse tasks such as object detection, semantic segmentation, panoptic segmentation, video processing, and few-shot classification. Vision Transformer (ViT) \cite{dosovitskiy2020image} presents a convolution-free transformer model for image classification. ViT processes input images as sequences of patch tokens, revolutionizing the traditional approach. Although ViT necessitates training on large-scale datasets to achieve optimal performance, DeiT \cite{touvron2021training} proposes a token-based distillation strategy that leverages a CNN as a teacher model, resulting in a competitive vision transformer trained on the ImageNet-1k dataset \cite{deng2009imagenet}. Building upon this foundation, concurrent research has extended transformer-based models to various domains, including video classification \cite{arnab2021vivit,bertasius2021space} and semantic segmentation 
\cite{liu2021swin,zheng2021rethinking}. SETR \cite{zheng2021rethinking} combines a ViT backbone with a standard CNN decoder. In contrast, Swin Transformer \cite{liu2021swin} adopts a variant of ViT that employs local windows shifted across layers. These advancements further demonstrate the versatility and effectiveness of Transformer-based architectures in computer vision tasks.

In this work, we present P2AT, a novel hybrid CNN-Transformer based encoder-decoder architecture, designed for semantic image segmentation. Our method leverages the powerful CNN backbone and integrates a Pyramid Pooling Axial Transformer as a global context information aggregator. Through extensive evaluations of widely adopted image segmentation benchmarks, our proposed method demonstrates competitive performance, showcasing its efficacy and potential for advancing real-time semantic image segmentation.

\section{Methodology}
\label{sec:methodology}
\subsection{Overall Architecture of P2AT}
\label{subsec:p2at_architecture}
Figure \ref{p2at} illustrates a diagram of the proposed P2AT. First, we present an overview of the proposed P2AT for real-time semantic image segmentation. Then, we analyze in detail the importance of several key elements that construct the model, including: (a) an encoder based on pre-trained  ResNet \cite{he2016deep} (b) the pyramid pooling axial attention which is one of the main building blocks of the method (c) the bidirectional fusion module which is used to fuse features of different stages efficiently, (d) decoder block.\\
Given an input image $I\in \mathbb{R}^{H\times W\times C}$  with channels C and spatial resolution W, H, we first utilize ResNet[36] to generate high-level features, and then integrate the proposed Transformer layers to complement the CNN on modeling the contextual features. Then, fed these features to a decoder that is introduced to maintain the global contextual information. After that, these high semantic features are fused with the features from low-level features through the bidirectional fusion module; BiF is efficient at combining features of different semantics. Finally, we enhanced the output feature using a global context enhancer module and refined them before the final prediction.

\subsection{Global and Local Feature Importance}
\label{sec:global_feature_importance}
A typical encoder-decoder framework usually utilizes the shallow layers to encode the high-resolution feature maps that carry target object detail information and the deeper layers to encode the higher semantics. However, simple upsampling strategies such as bilinear interpolation and deconvolution are incapable of collecting global context and restoring the missing information during the downsampling process. In this part of our work, we aim to improve semantic segmentation performance by designing a network capable of overcoming some of the problems of the encoder-decoder architecture. For this purpose, several modules and blocks have been developed and combined to construct the P2AT.  

\subsection{Bidirectional Fusion Module}
\label{sec:bidirectional_fusion}
\begin{figure*}
	\centering
	\includegraphics[width=0.7\linewidth]{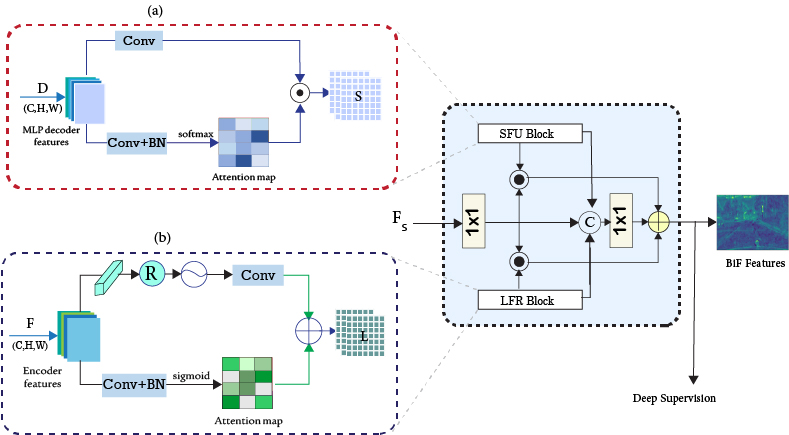}
	\caption{Efficient Bidirectional fusion Module.}
	\label{bifusion}
\end{figure*}

To efficiently combine the encoded features representations from a low-level encoder and high semantic features of the decoding module, we propose a new bidirectional fusion module (Figure \ref{bifusion}) that integrates both channel attention which used to transform the low-level features through the local feature refinement block, a semantic feature injection, and multi-stage multi-level fusion mechanisms. In equation \ref{bifusion}, $F_{BiF}$ is the multi-level fusion function.

\begin{equation}\label{eq:bifusion}
 B = F_{BiF}(D,L,F_{s})
\end{equation} 

Where D is the semantic descriptor, L denotes the detailed object features, and $F_{s}$ is the feature of the stage.

\textbf{Semantic Feature Upsampler (SFU) block}: SFU is proposed to gather semantic features, as shown in Figure \ref{bifusion}.(a). $D\in \mathbb{R}^{H\times W\times C}$ denote the output of layer $D_{5}$, and $D_{4}$ of the decoder. Feature Pyramid Network \cite{lin2017feature} is a simple architecture to propagate semantic features into rich detail features in the lower levels. By fusing semantic features with multi-scale features, the performance has been improved substantially in object detection and semantic segmentation. However, the process of reducing channels over stages causes a loss of important information. This paper introduces the semantic feature upsampler, a simple yet efficient upsampling that uses an attention mechanism to selectively inject global features into the BiF module. We formulated the semantic feature upsampler $S\in \mathbb{R}^{H\times W\times C}$ as follows: 

\begin{equation}\label{eq:egca}
S = \alpha(D;W_{\alpha}) \odot softmax(\beta(D;W_{\beta})
\end{equation} 

Where different $1\times 1$ convolutional layers ($\alpha$ and $\beta$) are used to map the input D, $\odot$ denotes the Hadamard product. 

\textbf{Local Feature Refinement block} The proposed architecture \ref{p2at} employs a bidirectional fusion design to facilitate the flow of information from different stages during the training process. In order to maintain the fusion of consistent semantic features, we introduce channel attention  (Figure \ref{bifusion}.b) to gather global information. In parallel to that, a spatial filter is integrated to suppress irrelevant information and enhance local details, as low-level encoder features
could be noisy. the local feature refinement block is formulated as follows:

\begin{equation}\label{eq:egca}
F_{g} = sigmoid(\eta(F;W_{\eta})
\end{equation} 

Where $\eta$ refers to a convolutional layer with kernels of $1\times 1$,

\begin{equation}\label{eq:egca}
L = \propto(F_{g} \oplus \gamma(G(F);W_{\gamma});W_{\propto})
\end{equation} 
 
Where ($\gamma$, $\propto$) represent convolution with kernels of $1\times 1$, G denotes the global average pooling. 

\textbf{Global Context Enhancer (GCE)} is introduced to compensate for the inadequacy of concatenating different semantic levels. Given the input feature $F_BiF$, the global context enhancer module first applies global average pooling to gather global semantic information
on and uses a gating mechanism to selectively choose the informative high semantic descriptors, which help remove the noise that can be introduced into the BiF by the shallower stages of the encoder. 

\begin{figure}[!h]
	\includegraphics[width=0.9\linewidth]{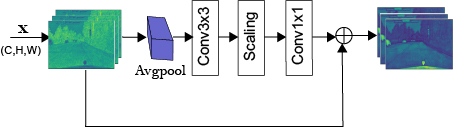}
	\caption{Global Context Enhancer Module.}
	\label{Fig:global_context}
\end{figure}

\subsection{Feature Decoding and Refinement}
The decoder block (refer to Figure.\ref{Fig:global_context}.(b))constructs of depth-wise convolution with kernel sizes of 3, 5 and 7 for stages 5, 4, and 3, respectively, followed by batch normalization. Then, we employ two point-wise convolution layers to enrich the local representation and help maintain object context. Unlike ConvNeXt \cite{liu2022convnet}, which has used Layer Normalization and Gaussian error Unit activation, we have used Hardswish activation \cite{howard2019searching} for non-linear feature mapping. Finally, a skip connection is added to facilitate the information flow across the network hierarchy. This decoder can be represented as follows:

\begin{figure}[!h]
    \centering
	\includegraphics[width=1\linewidth]{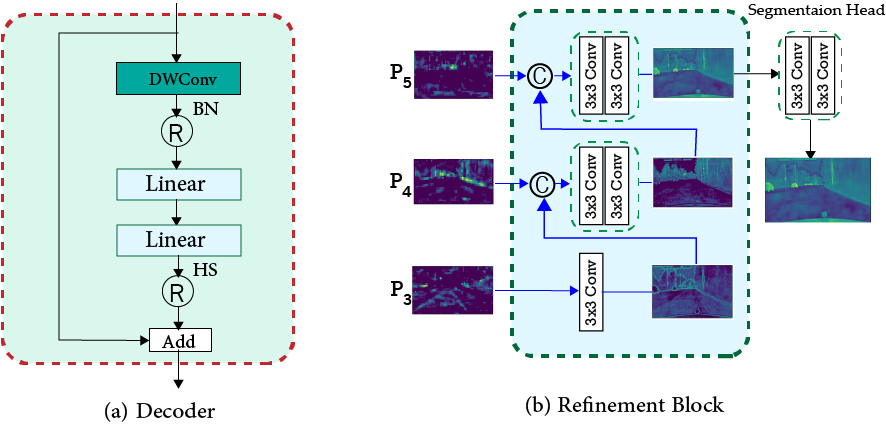}
	\caption{Illustrates the detail of the (a) feature refinement module and (b) feature decoding block.}
	\label{Fig:refinement_module}
\end{figure} 

\begin{equation}
D_{i+1} = D_{i} + f_{L}^{H}(f_{L}(f^{DW}_{k\times k}(D_{i})))
\end{equation}
where $D_{i}$ is an input feature maps of shape  $H\times W\times C$, $f_{L}^{H}$ denotes the point-wise convolution layer followed by Hardswish, $f^{DW}$ is a depth-wise convolution with kernel size of $k\times k$, and $D_{i+1}$ denotes the output feature maps of the decoder block.

The refinement block (Figure. \ref{Fig:refinement_module}.b is introduced to filter the noisy features that produced by the decoder more accurate per-pixel classification and localization.
\begin{figure}[h!]
	 \centering
	\includegraphics[width=1\linewidth]{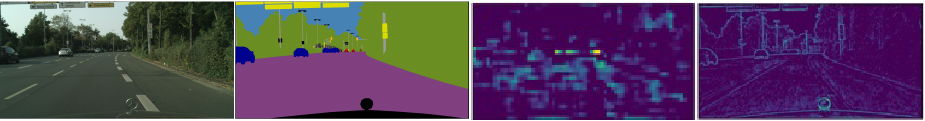}
	\caption{Feature visualization of the refinement module. Left to right are original image, ground truth, output of decoder block, and the maps of the refined feature.}
	\label{Fig:refinement_visualization}
\end{figure}

\subsection{Scale-aware Semantic Aggregation Block}
The scale-aware semantic aggregator consists of L number of stacked Pyramid Pooling Axial Transformer blocks. Each Transformer comprises a pyramid pooling axial attention module and Feed-Forward Network (FFN). The output of the $l-th$ $(l\in [1,2,..,L])$ Axial Transformer blocks can be presented as follows:

\begin{equation}
    \acute{z}_{l} = P2A2(z_{l-1}) + z_{l-1}
\end{equation}

\begin{equation}
  z_{l} = MLP(\acute{z}_{l})+ \acute{z}_{l}
\end{equation}
 Where $z$, $z_{l} $, and $z_{l}$ denote the input, output of the axial attention and the output of the Transformer block, respectively. P2A2 is the abbreviation of Pyramid Pooling Axial Attention. 
 
\begin{figure*}
    \begin{center}
	\includegraphics[width=0.8\linewidth]{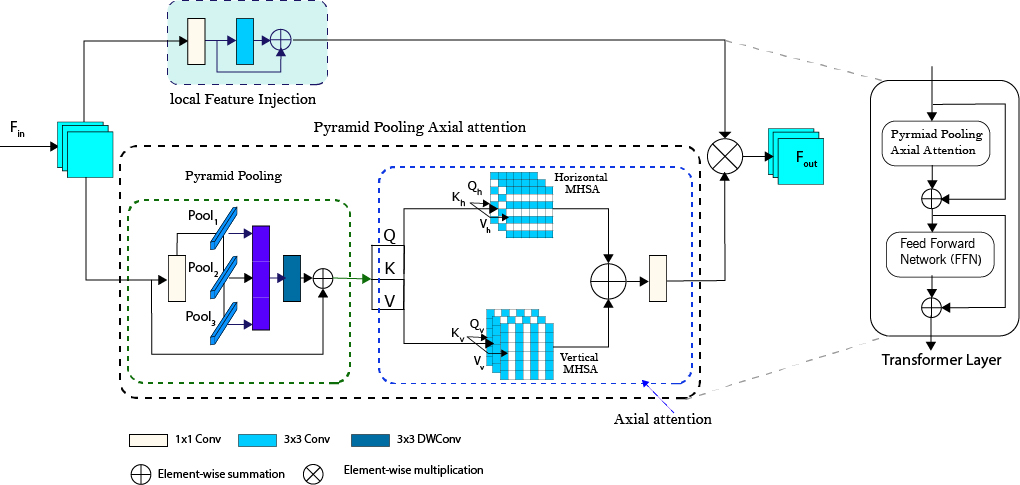}
	\caption{Right: Illustration of the proposed Scale-aware Semantic Aggregation Block including a Pyramid Pooling Axial attention and a Feed-Forward Network (FFN). Left is the
    Pyramid Pooling Axial Attention, Pyramid Pooling and Axial Attention.}
	\label{Fig:transformer_block}    
    \end{center}
\end{figure*}

\textbf{Pyramid Pooling Axial Attention}
Here we present the proposed pyramid pooling axial attention. The main structure is illustrated in Figure \ref{Fig:transformer_block}. First, the input features are fed into a pyramid pooling submodule, which captures global context information by performing pooling operations with different kernel sizes, Equation \ref{pa:p}. Next, the axial attention module is applied to the pooled features to capture spatial dependencies in both the vertical and horizontal axes. This module leverages positional embedding \cite{wan2023seaformer} to encode spatial information and generate attention maps that highlight relevant regions in the image. The attention maps are then combined with residual blocks to enhance the spatial details, allowing the model to generate informative contextual information that enables our architecture to obtain high accuracy on a very small dataset.
\begin{equation}
   X = \theta(F_{in};W_{\theta})
\end{equation}
\begin{equation}\label{pa:p1}
    \textbf{P}_{1} = AvgPool_{3}(X) 
\end{equation}
\begin{equation}\label{pa:p2}
    \textbf{P}_{2} = AvgPool_{5}(\textbf{P}_{1}) 
\end{equation}
\begin{equation}\label{pa:p3}
    \textbf{P}_{3} = AvgPool_{7}(\textbf{P}_{2}) 
\end{equation}
Where $\textbf{P}_{3}$, $\textbf{P}_{5}$ , $\textbf{P}_{7}$ is the generated pyramid features. Next, we sum up the pyramid feature maps and employ a convolutional operation on top of it.
\begin{equation}\label{pa:p}
    \textbf{P}^{pp} = W_{3,3}^{DW}([P_{3},P_{5},P_{7}]) 
\end{equation}

Axial Attention \cite{ho2019axial,wang2020axial} has been introduced to reduce the computation cost of the attention network. Since then, it has been integrated in many semantic segmentation frameworks \cite{lou2022caranet,huang2022channelized}.\\

\begin{table*}[hpt]
	\caption{The Per-class, class, and category IoU evaluation on the Cityscapes Test set. List of classes from left to right: Road, Sidewalk, Building, wall, Fence, Pole, Traffic light, Traffic sign, Vegetation, Terrain, Sky, Pedestrian, Rider, Car, Truck, Bus, Train, Motorbike, and Bicycle."cla->mIoU". ”-” indicates the corresponding result is not reported by the methods}
	\label{Table:Tab8}
	\vspace{1ex}
	\begin{center}
		\begin{adjustbox}{width=1\textwidth}
			\small
			\begin{tabular}{lcccccccccccccccccccc}
				\hline 				Method&Road&Sidewalk&Building&Wall&Fence&Pole&Traffic light&Traffic sign&Vegetation&Terrain&Sky&Person&Rider&Car&Truck&Bus&Train&Motor&Bicyclist&mIoU\\
				\hline
				\hline
			CRF-RNN\cite{zheng2015conditional}&96.3&73.9&88.2&47.6&41.3&35.2&49.5&59.7&90.6&66.1&93.5&70.4&34.7&90.1&39.2&57.5&55.4&43.9&54.6&62.5\\
            FCN\cite{long2015fully}&97.4&78.4&89.2&34.9&44.2&47.4&60.1.5&65.0&91.4&69.3&93.9&77.1&51.4&92.6&35.3&48.6&46.5&51.6&66.8&65.3\\
			DeepLabv2\cite{chen2017deeplab}&-&-&-&-&-&-&-&-&-& -&-&-&-&-&-&-&-&-&-&70.4\\              Dilation10\cite{yu2016multi}&97.6&79.2&89.9&37.3&47.6&53.2&58.6&65.2&91.8&69.4&93.7&78.9&55.0&93.3&45.5&53.4&47.7&52.2&66.0&67.1\\
			AGLNet\cite{zhou2020aglnet}&97.8& 80.1& 91.0&51.3&50.6&58.3&63.0&68.5&92.3&71.3&94.2&80.1&59.6&93.8&48.4&68.1&42.1&52.4& 67.8&70.1\\
			BiSeNetV2\slash BiSeNetV2\_L\cite{yu2021bisenet}&-&-&-&-&-&-&-&-&-& -&-&-&-&-&-&-&-&-&-&73.2\\
			LBN-AA\cite{dong2020real}&98.2&84.0&91.6&50.7&49.5&60.9&69.0&73.6&92.6& 70.3&94.4&83.0&65.7&94.9&62.0&70.9&53.3&62.5&71.8-&73.6\\
			\hline
			P2AT-S &98.4&84.4&92.9&54.4&56.6&\textbf{67.7}&\textbf{75.2}&\textbf{78.1}&\textbf{93.5}&71.9&\textbf{95.6}&\textbf{85.7}&69.2&95.5&62.2&81.1&76.3&64.7&74.9&77.8 \\
			P2AT-M &\textbf{98.5}&\textbf{85.7}&\textbf{93.0}&\textbf{58.6}&\textbf{58.3}&67.3&74.7&77.7&\textbf{93.5}&\textbf{72.2}&\textbf{95.6}&85.3&69.2&95.6&\textbf{68.1}&83.0&\textbf{78.3}&65.9&\textbf{75.1}&\textbf{78.7}\\
            P2AT-L&\textbf{98.5}&85.3&92.6&53.1&57.4&66.6&74.7&\textbf{78.1}&93.3&69.8&95.1&86.0&\textbf{70.5}&\textbf{95.8}&67.8&\textbf{83.3}&	72.2&\textbf{67.0}&74.8&78.0\\
			\hline
			\end{tabular}
		\end{adjustbox}
	\end{center}
\end{table*}  

\section{Experimental Results}
In this section, we present the results of comprehensive experiments that were conducted to evaluate the effectiveness of P2AT on three challenging benchmarks: Camvid \cite{brostow2009semantic}, Cityscapes \cite{cordts2016cityscapes}, and PASCAL VOC 2012 \cite{everingham2010pascal}. We first present the datasets and implementation details. Then, the ablation studies performed to ensure the validity of each component in P2AT. Finally, we compare our method with other state-of-the-art networks using the standards metrics: number of trainable parameters, floating point operations (flops), and class mean intersection over union (mIoU)
\subsection{Experiments Settings}
\subsubsection{Camvid} Camvid is a challenging road scene understanding dataset consisting of 376 training images, 101 validation images, and 233 test images. The dataset is small, with only 376 training images, and the distribution of labels is unbalanced. This makes it difficult for models to learn to segment images accurately.
To make a fair comparison with other state-of-the-art models, we evaluated our model on 11 classes, including building, sky, tree, car, and road. The class 12th was marked as ignore class to hold the unlabelled data..\\
\subsubsection{Cityscapes} The Cityscapes dataset is a benchmark for urban scene understanding. It contains 5000 high-resolution images of 2048x1024 pixels with fine annotations captured from different cities. The dataset is divided into 2975 training images, 500 validation images, and 1525 test images. We evaluated our model on 19 semantic segmentation classes, including road, sidewalk, building, vegetation, and sky. Despite the challenges, the Cityscapes dataset is a valuable resource for researchers and practitioners working on urban scene understanding. The dataset provides a large and diverse set of images that can be used to train and evaluate models.

\subsubsection{Implementation Details}
Our training settings are closely aligned with previous works 
 \cite{chen2017deeplab,zhao2017pyramid}. Specifically, we have employed SGD optimizer with a poly-learning rate strategy. Additionally, we have incorporated various data augmentation techniques that have been used by other methods for a fair comparison, including random horizontal flipping, random cropping, and random scaling within the range of 0.5 to 2.0. For the Cityscapes, Camvid, and PASCAL VOC datasets, we have established the following key training parameters: 500 epochs, an initial learning rate of $1e-2$, weight decay of $5e-4$, cropped image size of $1024\times 1024$, and a batch size of 12 for Cityscapes; 140 epochs, an initial learning rate of $5e-3$, weight decay of $5e-4$, cropped image size of $960\times 720$, and a batch size of 12 for Camvid; and 200 epochs, an initial learning rate of $1e-3$, weight decay of $1e-4$, cropped image size of $512\times 512$, and a batch size of 16 for PASCAL VOC.
During the inference stage, prior to testing, our models underwent training using both the training and validation sets for the Cityscapes and Camvid datasets. To evaluate the inference speed, we conducted measurements using a platform equipped with a single RTX 3090 GPU, PyTorch \cite{pytorch} 1.8, CUDA 11.7, cuDNN 8.0, and an Ubuntu environment. We conducted thorough evaluations of the inference speed to ensure the robustness and validity of our results.
\begin{figure*}
    \begin{center}
    \begin{subfigure}{0.18\textwidth}
		\includegraphics[width=3.5cm,height=1.5cm]{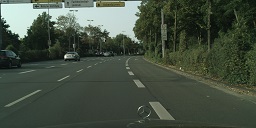}
	\end{subfigure}
	\begin{subfigure}{0.18\textwidth}
		\includegraphics[width=3.5cm,height=1.5cm]{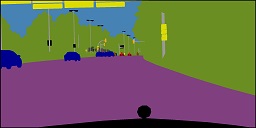}
	\end{subfigure}
	\begin{subfigure}{0.18\textwidth}
		\includegraphics[width=3.5cm,height=1.5cm]{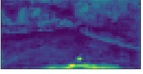}
	\end{subfigure}
	\begin{subfigure}{0.18\textwidth}
		\includegraphics[width=3.5cm,height=1.5cm]{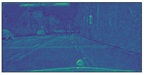}
	\end{subfigure}
        \begin{subfigure}{0.18\textwidth}
		\includegraphics[width=3.5cm,height=1.5cm]{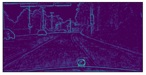}
	\end{subfigure}
         \\
         \begin{subfigure}{0.18\textwidth}
		\includegraphics[width=3.4cm,height=1.5cm]{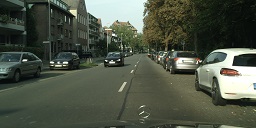}
	\end{subfigure}
	\begin{subfigure}{0.18\textwidth}
		\includegraphics[width=3.4cm,height=1.5cm]{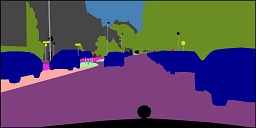}
	\end{subfigure}
	\begin{subfigure}{0.18\textwidth}
		\includegraphics[width=3.4cm,height=1.5cm]{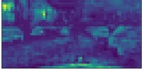}
	\end{subfigure}
	\begin{subfigure}{0.18\textwidth}
		\includegraphics[width=3.4cm,height=1.5cm]{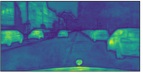}
	\end{subfigure}
        \begin{subfigure}{0.18\textwidth}
		\includegraphics[width=3.4cm,height=1.5cm]{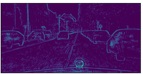}
	\end{subfigure}
        \\
     \begin{subfigure}{0.18\textwidth}
		\includegraphics[width=3.4cm,height=1.5cm]{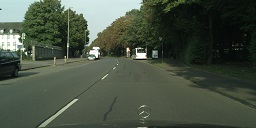}
	\end{subfigure}
	\begin{subfigure}{0.18\textwidth}
		\includegraphics[width=3.4cm,height=1.5cm]{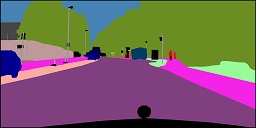}
	\end{subfigure}
	\begin{subfigure}{0.18\textwidth}
		\includegraphics[width=3.4cm,height=1.5cm]{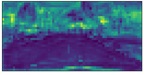}
	\end{subfigure}
	\begin{subfigure}{0.18\textwidth}
		\includegraphics[width=3.4cm,height=1.5cm]{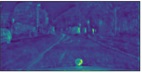}
	\end{subfigure}
        \begin{subfigure}{0.18\textwidth}
		\includegraphics[width=3.4cm,height=1.5cm]{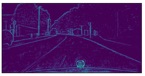}
	\end{subfigure}
    \\
     \begin{subfigure}{0.18\textwidth}
		\includegraphics[width=3.4cm,height=1.5cm]{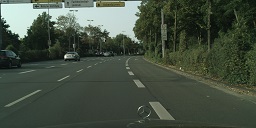}
	\end{subfigure}
	\begin{subfigure}{0.18\textwidth}
		\includegraphics[width=3.4cm,height=1.5cm]{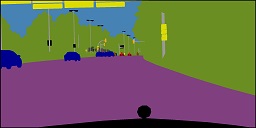}
	\end{subfigure}
	\begin{subfigure}{0.18\textwidth}
		\includegraphics[width=3.4cm,height=1.5cm]{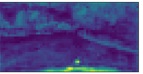}
	\end{subfigure}
	\begin{subfigure}{0.18\textwidth}
		\includegraphics[width=3.4cm,height=1.5cm]{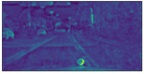}
	\end{subfigure}
        \begin{subfigure}{0.18\textwidth}
		\includegraphics[width=3.4cm,height=1.5cm]{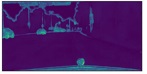}
	\end{subfigure}
	\caption{Feature maps visualization of our proposed architecture on the Cityscapes validation dataset. From left to right: (a) original input images; (b) The ground truth; (c) Scale-aware Semantic Aggregation block ; (d) bidirectional fusion (stage4) maps; (e) global context enhancer (stage 4); (f) the feature maps of the segmentation head.}
	\label{cityscapes_fig}
 \end{center}
\end{figure*}

\subsection{Ablation Study}
We perform ablation studies on Camvid and Pascal VOC2012 datasets to evaluate the impact of each component in the proposed P2AT. The results on Camvid reported on the testing set, while Pascal VOC2012 results reported on the validation set.
			
\subsection{Results on Pascal VOC2012}

PASCAL VOC 2012 \cite{everingham2010pascal} consists of images with different resolutions, representing 21 classes, including the background class. The dataset comprised of 1,464 training, 1,449 validation, and 1,456 test images. The training set was extended to a total of 10,582 images \cite{hariharan2011semantic}. While this dataset is not commonly used for evaluating real-time segmentation methods, its low-resolution images can be utilized to run ablation studies, making it suitable for initial tests.
Table \ref{tab:pascal2012_val} presents backbone, accuracy, flops, and parameters for our model compared to existing work. We specifically chose methods that reported results on the PASCAL VOC validation set without employing inference strategies such as horizontal mirroring and multi-scale testing. The results indicate that our medium size of our method P2AT-M achieves competitive accuracy of 79.6\%mIoU 

\begin{table}
    \centering
	\caption{\MakeUppercase{Performance comparison of P2AT variants with state-of-the-art methods on Pascal VOC 2012 Validation set.}}
	\label{tab:pascal2012_val}
	\vspace{1ex}
	\begin{adjustbox}{width=0.48\textwidth}
			\small
		\begin{tabular}{lcccc}
			\hline
			Method&Backbone&Params&Flops&mIoU(\%)\\ 
			\hline
			\hline
			DeepLabV3\cite{chen2017rethinking}&ResNet-101& 58.6& 249.2& 78.5\\
			Auto-DeepLab-L\cite{liu2019auto}&-&44.42&79.3&73.6\\
			DeepLabV3Plus\cite{chen2018encoder}&Xception-71&43.48&177&80.0\\
			DFN\cite{yu2018learning}&ResNet-101&-&-&79.7\\
			SDN\cite{fu2019stacked}&DenseNet-161&238.5&-&79.9\\
			HyperSeg-L\cite{nirkin2021hyperseg}&EfficientNet-B1&39.6&8.21&80.6\\
			\hline
            P2AT-M&ResNet-34&41.7&37.5&79.6\\
			\hline
		\end{tabular}
	\end{adjustbox}
\end{table}

\subsection{Ablation on Camvid and Cityscapes}
In this subsection, we conduct a series of experiments to explore different variants of our proposed method. 
\begin{table}
	\caption{\MakeUppercase{Comparison between P2AT-S, P2AT-M, and P2AT-L in terms of parameters, flops, frames per second (FPS), and accuracy on Cityscapes and Camvid datasets}}
	\label{table:camvid_city_ablation}
	\centering
	\begin{adjustbox}{width=\linewidth}
		\small
		\begin{tabular}{ccccccccc}
			\hline  
			\multirow{2}{*}{Method} & \multirow{2}{*}{Backbone} & \multirow{2}{*}{Params} & \multicolumn{3}{c}{Cityscapes} & \multicolumn{3}{c}{Camvid} \\ 
			\cline{4-6} \cline{7-9}
			&&& Flops & FPS & mIoU \% & Flops & FPS & mIoU \% \\
			\hline\hline
			P2AT-S & ResNet-18 & 12.6 & 80.2 & 80.1 & 78.0 & 73 & 113.6 & 80.5 \\
			P2AT-M & ResNet-34 & 22.7 & 118 & 61.8 & 79.8 & 96 & 86.6 & 81.0 \\
			P2AT-L & ResNet-50 & 37.5 & 166 & 40.6 & 78.8 & 112 & 56.1 & 81.1 \\
			\hline
		\end{tabular}
	\end{adjustbox}
\end{table}
Table \ref{table:camvid_city_ablation} presents the results of the ablation study on variants of P2AT models for semantic segmentation in Camvid and Cityscapes datasets. P2AT-S has a more compact model architecture than the other variants, with only 0.8\% mIoU less than P2AT-L and 1.8 mIoU than the variant with the highest accuracy P2AT-M. The different settings of P2AT have achieved very high performance on the Camvid test set and competitive results on the cityscapes and Pascal VOC2012 datasets. 

Model complexity vs. performance: As the model complexity increases (from P2AT-S to P2AT-L), the number of parameters and flops increases. However, the impact on performance varies. P2AT-M achieves higher mIoU than P2AT-S, indicating a better trade-off between model complexity and accuracy. P2AT-L, despite having higher parameters and flops, achieves a slightly lower mIoU than P2AT-M, suggesting diminishing returns in performance improvement with increased model complexity or needing more optimization to utilize the whole parameters. Overall, the ablation study provides insights into the trade-offs between model complexity, computational efficiency, and accuracy in semantic segmentation for road scene understanding

\subsection{Results on Cityscapes Dataset}
Table \ref{miou_cityscapes} provides a comprehensive comparison between three variants of P2AT (P2AT-S, P2AT-M, P2AT-L) and other state-of-the-art (SOTA) approaches on the Cityscapes test dataset. The comparison includes various aspects such as backbone architecture, input resolution, GPU type, number of parameters, flops (floating-point operations), evaluation split(val, test), achieved accuracy (mIoU), and inference speed (FPS).
Firstly, it is evident that the proposed P2AT-S, P2AT-M, and P2AT-L models obtain a competitive result on the Cityscapes dataset. P2AT-S, based on ResNet-18, achieves an accuracy of 77.8\% mIoU on the test set, outperforming many existing methods such as SwiftNetRN-18, SFNet (DF1), FANet-18, LBN-AA, and AGLNet. P2AT-M, using ResNet-34, further improves the performance, obtaining an accuracy of 78.7\% mIoU, surpassing several methods, including transformer-based methods such as SeaFormer-S, and SegFormer, as well as other plane methods, PP-LiteSeg-T2, and HyperSeg-M. P2AT-L, utilizing ResNet50, achieves a mIoU of 78\%, comparable to other top-performing methods like HyperSeg-S and BiSeNetV1.
Compare to the SOTA methods, it is notable that our proposed P2AT models demonstrate competitive accuracy while maintaining reasonable computational efficiency. P2AT-S achieves comparable accuracy to SegFormer, SwiftNetRN-18, and SFNet (DF1), while offering faster inference speeds. P2AT-M exhibits improved accuracy compared to SwiftNetRN-18, PP-LiteSeg-T2, and HyperSeg-M, while maintaining a comparable inference speed. P2AT-L achieves competitive accuracy similar to HyperSeg-S and BiSeNetV1, although at a slightly lower inference speed.
Additionally, the proposed P2AT models exhibit favorable trade-offs between accuracy and efficiency compared to other existing methods. For instance, P2AT-M achieves a higher mIoU than SeaFormer-S, SeaFormer-B, and SwiftNetRN-18 ens, while being more computationally efficient in terms of parameters and flops. Similarly, P2AT-L achieves comparable accuracy to HyperSeg-S and BiSeNetV1 with a significantly lower number of parameters and flops. These observations highlight the effectiveness of the proposed models in achieving competitive performance while maintaining computational efficiency.
Furthermore, it is important to note that the proposed P2AT models leverage different backbone architectures (ResNet-18, ResNet-34, and ResNet-50). The use of deeper backbones (P2AT-M and P2AT-L) contributes to improved performance, as evident from their higher mIoU values compared to P2AT-S. This demonstrates the importance of feature representation capacity in achieving better accuracy, but a well optimized training is needed to benefit from the larger backbones.
In terms of input resolution, all P2AT models adopt a fixed resolution of 1024x1024. Despite using a relatively lower resolution compared to some methods like SwiftNetRN-18 (1024x2048) and SFNet (DF1) (1024x2048), the proposed models achieve competitive accuracy. This suggests that the integration of pyramid pooling axial attention, and proposed feature fusion modules in the P2AT architecture enables effective multi-scale feature extraction, compensating for the lower input resolution.
In conclusion, the proposed P2AT variants exhibit competitive performance on the Cityscapes dataset compared to other SOTA methods. They achieve high accuracy while maintaining reasonable computational efficiency.

\begin{table*}
	\caption{\MakeUppercase{Comparison between the proposed method S\textsuperscript{2}-FPN and the other SOTA methods on the Cityscapes test dataset. We report the backbone, input resolution, GPU type, number of parameters (M), flops (G), evaluation split (set), achieved accuracy (mIoU), and the inference speed (FPS)}}
	\label{miou_cityscapes}
	\centering
		\begin{adjustbox}{width=1\textwidth}
			\small
			\begin{tabular}{ccccccccc}
				\hline
				\multirow{2}{*}{Method}&\multirow{2}{*}{Backbone}&\multirow{2}{*}{Resolution}&\multirow{2}{*}{GPU}&\multirow{2}{*}{\#Params}&\multirow{2}{*}{\#GFLOPs}&\multirow{2}{*}{\#FPS}& \multicolumn{2}{c}{mIoU}\\ 
                \cline{8-9}
               &&&&&&&Val&Test\\
				\hline\hline
				SeaFormer-S\cite{wan2023seaformer}&SeaFormer-S&1024$\times$1024&GTX 1080Ti &-&8.0&-&76.1&75.9\\
				SeaFormer-B\cite{wan2023seaformer}&SeaFormer-B&1024$\times$1024&GTX 1080Ti &-&13.7&&77.7&77.5\\
    			SegFormer&MiT-B0&1024$\times$1024&Tesla V100 &\textbf{3.8}&125.5&15.2&-&76.2\\
				\hline
                SwiftNetRN-18\cite{orsic2019defense}&ResNet-18&1024$\times$2048&GTX 1080Ti&11.8&104.0& 39.9&75.5&75.4 \\
				SwiftNetRN-18 ens\cite{orsic2019defense}&ResNet-18&1024$\times$2048&GTX 1080Ti&24.7&218.0&18.4&-&76.5\\
				\hline
                PP-LiteSeg-T2\cite{peng2022pp}&STDC1&768$\times$1536&GTX 1080Ti&-&-& 143.6&76.0&74.9\\
				PP-LiteSeg-B2\cite{peng2022pp}&STDC2&768$\times$1536&GTX 1080Ti&-&-&102.6&78.2&77.5\\
				\hline
				ENet\cite{zheng2015conditional}&No&640 $\times$360&TitanX&0.4&3.8& 135.4&-&58.3\\
				ICNet\cite{chen2014semantic}&PSPNet50&1024$\times$2048&TitanX&26.5&28.3&30.3&-&69.5\\
				DABNet\cite{long2015fully}&No &512$\times$1024& GTX 1080Ti&0.76&10.4&104&-&70.1\\
				DFANet-A\cite{li2019dfanet}&XceptionA&1024$\times$1024&Titan X&7.8&3.4&100&-&71.3\\
				DFANet-B\cite{li2019dfanet}&XceptionB&1024$\times$1024&Titan X&4.8&\textbf{2.1}&120&-&67.1\\
				FasterSeg\cite{chen2019fasterseg}&No &1024$\times$2048&GTX 1080Ti&-&-&163.9&&71.5\\
				TD4-Bise18\cite{hu2020temporally}&BiseNet18&1024$\times$2048&Titan Xp&-&-&-&-&74.9\\
				\hline
                SFNet(DF1)\cite{li2020semantic}&DF1&1024$\times$2048&GTX 1080Ti GPU&9.03&-&74&-&74.5\\
				SFNet(DF2)\cite{li2020semantic}&DF2&1024$\times$2048&GTX 1080Ti GPU&10.53&-&53.0&-&77.8\\
				\hline
				FANet-18\cite{hu2020real}&ResNet-18&1024$\times$2048&Titan X&-&49&72&&74.4\\
				FANet-34\cite{hu2020real}&ResNet-34&1024$\times$2048&Titan X&-&65&58&-&75.5\\
				LBN-AA\cite{dong2020real}&LBN-AA+MobileNetV2&488$\times$896&Titan X&6.2&49.5&51.0&-&73.6\\
				AGLNet\cite{zhou2020aglnet}&No&512$\times$1024& GTX 1080Ti&1.12&13.88&71.3&&52.0\\
				HMSeg\cite{li2020humans}&No&768$\times$1536& GTX 1080Ti&2.3&-&74.3 &-&83.2\\
				TinyHMSeg\cite{li2020humans}&No&768$\times$1536& GTX 1080Ti&0.7&-&71.4&-&172.4\\
                \hline
				BiSeNetV1\cite{yu2018bisenet} &ResNet-18&768$\times$1536& GTX 1080Ti&49.0&55.3&65.5&74.8&74.7\\
                BiSeNetV2\cite{yu2021bisenet}&No&512$\times$1024& GTX 1080Ti&-&-&156&73.4&72.6\\
				BiSeNetV2-L\cite{yu2021bisenet}&No&512$\times$1024& GTX 1080Ti&-&118.5&47.3&75.8&75.3\\
                \hline
				STDC1-Seg50\cite{fan2021rethinking}&STDC1&512$\times$1024&GTX 1080Ti &8.4&-&250.4 &72.2&71.9\\
				STDC2-Seg50\cite{fan2021rethinking}&STDC2&512$\times$1024&GTX 1080Ti &12.5&-&\textbf{188.6} &74.2&73.4\\
				STDC1-Seg75\cite{fan2021rethinking}&STDC1&768$\times$1536&GTX 1080Ti &8.4&-&126.7 &74.5&75.3\\
				STDC2-Seg75\cite{fan2021rethinking}&STDC2&768$\times$1536&GTX 1080Ti&12.5&-&97.0 &77.0&76.8\\
				\hline
                HyperSeg-S\cite{nirkin2021hyperseg}&EfficientNet-B1&768$\times$1536& GTX 1080Ti&10.2&17.0&16.1&78.2&78.1\\
				HyperSeg-M\cite{nirkin2021hyperseg}&EfficientNet-B1&512$\times$1024& GTX 1080Ti&10.1&7.5&36.9&76.2&75.8\\
				\hline
				\hline 
				P2AT-S&ResNet-18&1024$\times$1024&RTX 3090&12.6&80.2&80.1&78.0&77.8\\
				P2AT-M&ResNet-34&1024$\times$1024&RTX 3090&22.7&118&61.8&\textbf{79.8}&\textbf{78.7}\\
				P2AT-L&ResNet-50&1024$\times$1024&RTX 3090&73.5&166&40.6&78.8&78.0\\
				\hline
			\end{tabular}
		\end{adjustbox}
\end{table*}


\begin{figure*}
	\centering
	\begin{subfigure}{0.24\textwidth}
		\centering
		\includegraphics[width=4.5cm,height=1.4cm]{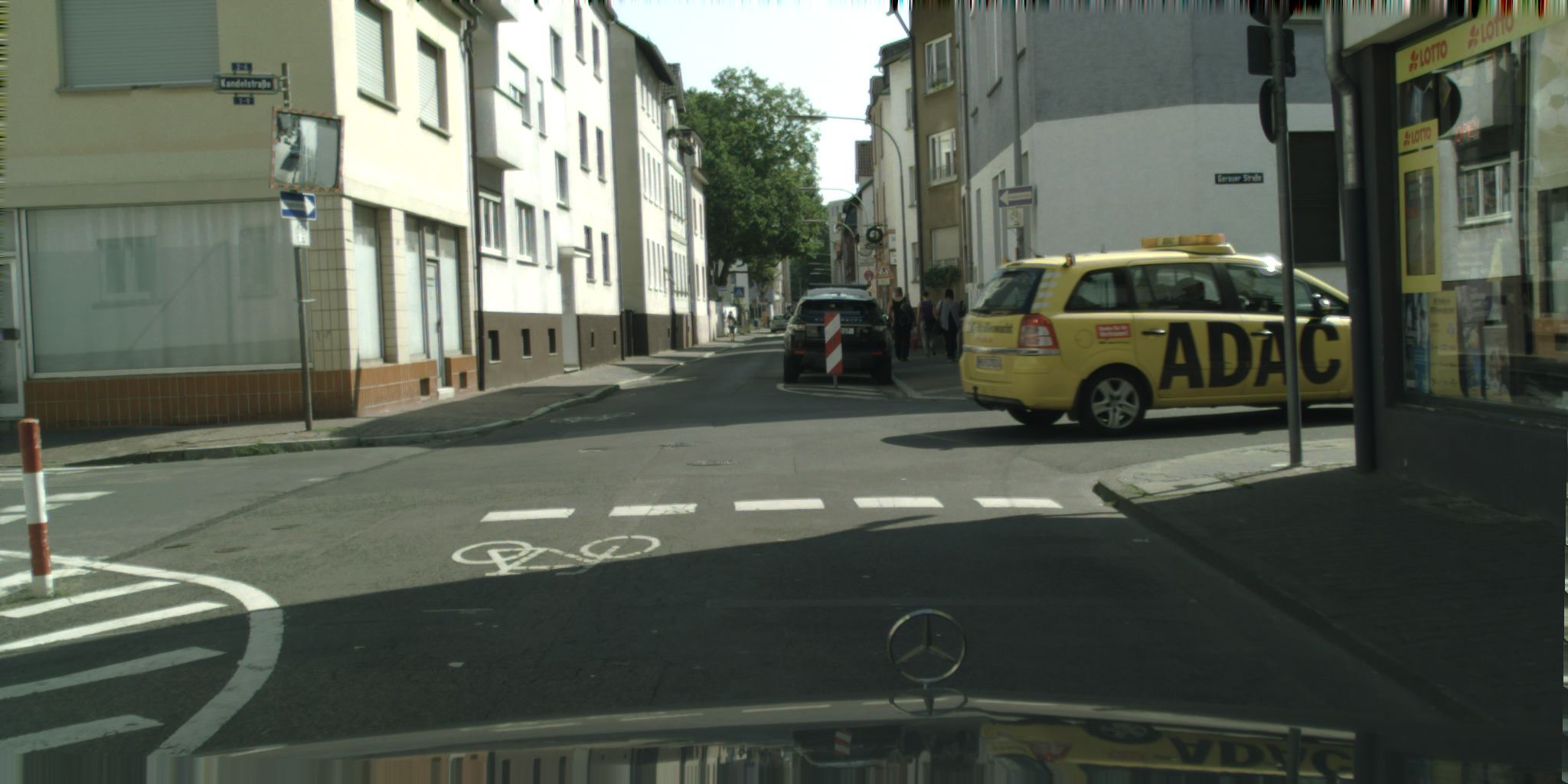}
	\end{subfigure}
	\begin{subfigure}{0.24\textwidth}
		\centering
		\includegraphics[width=4.8cm,height=1.4cm]{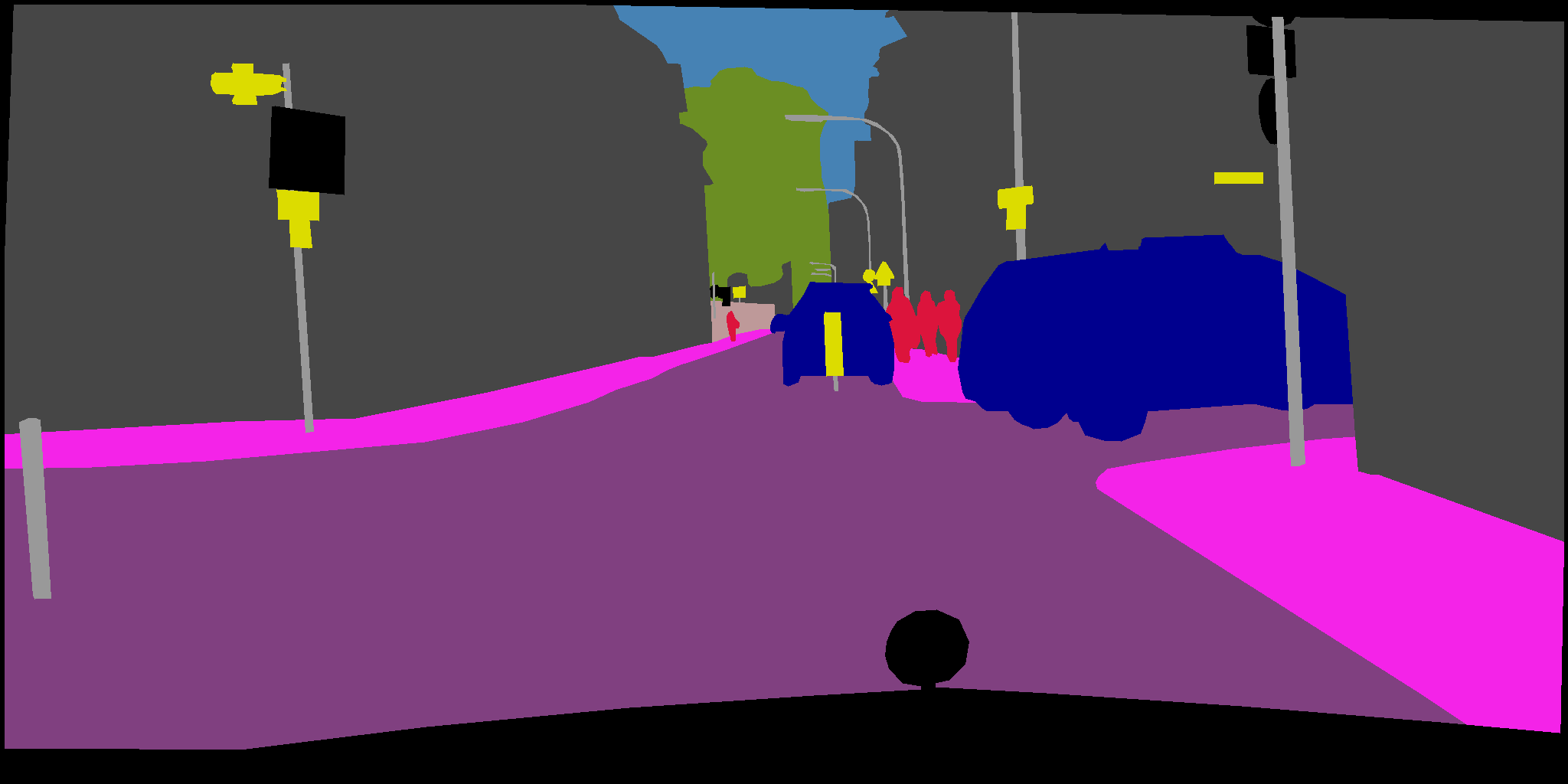}
	\end{subfigure}
	\begin{subfigure}{0.24\textwidth}
		\centering
		\includegraphics[width=4.8cm,height=1.4cm]{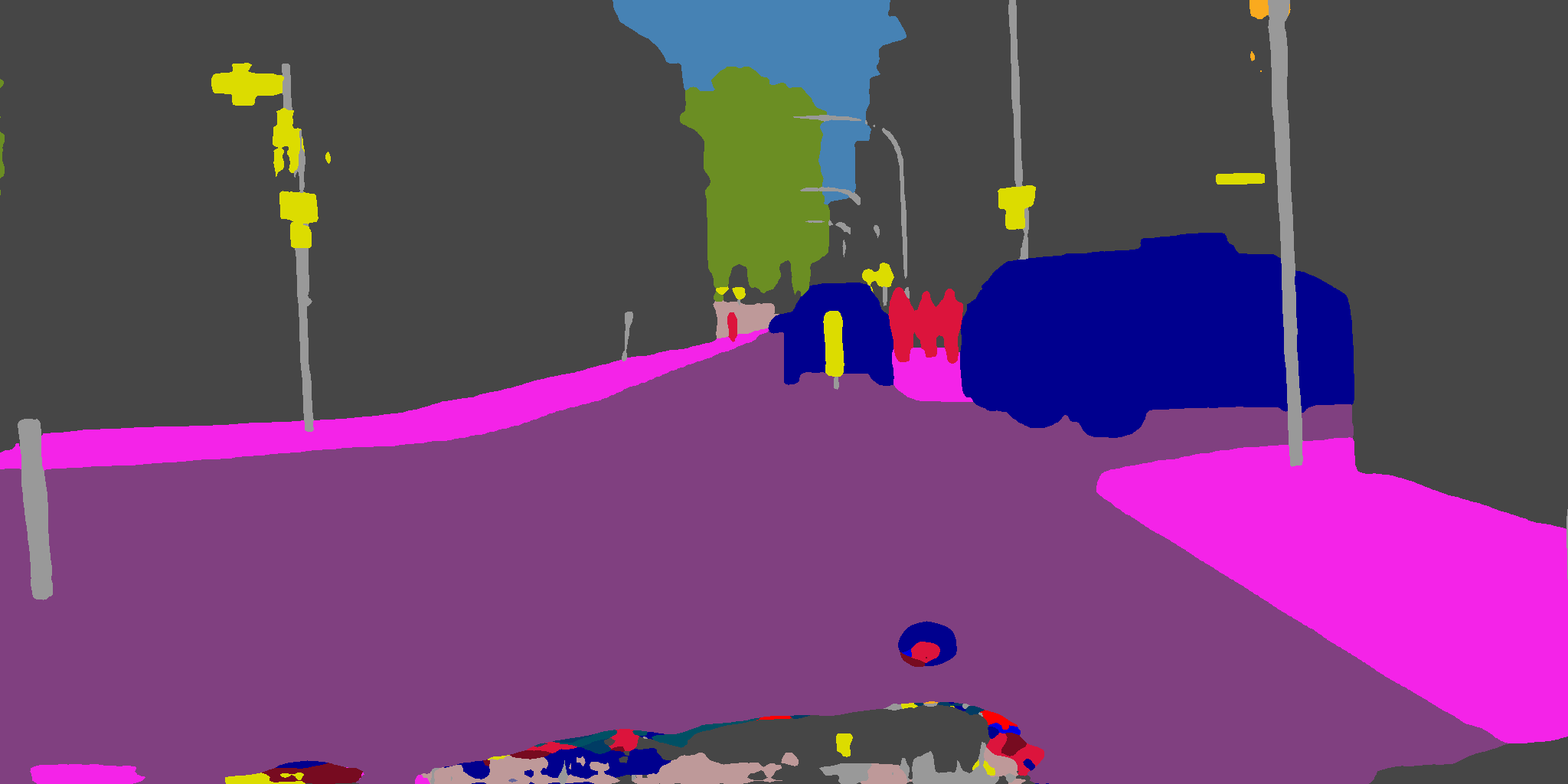}
	\end{subfigure}
	\\
	\begin{subfigure}{0.24\textwidth}
		\centering
		\includegraphics[width=4.8cm,height=1.4cm]{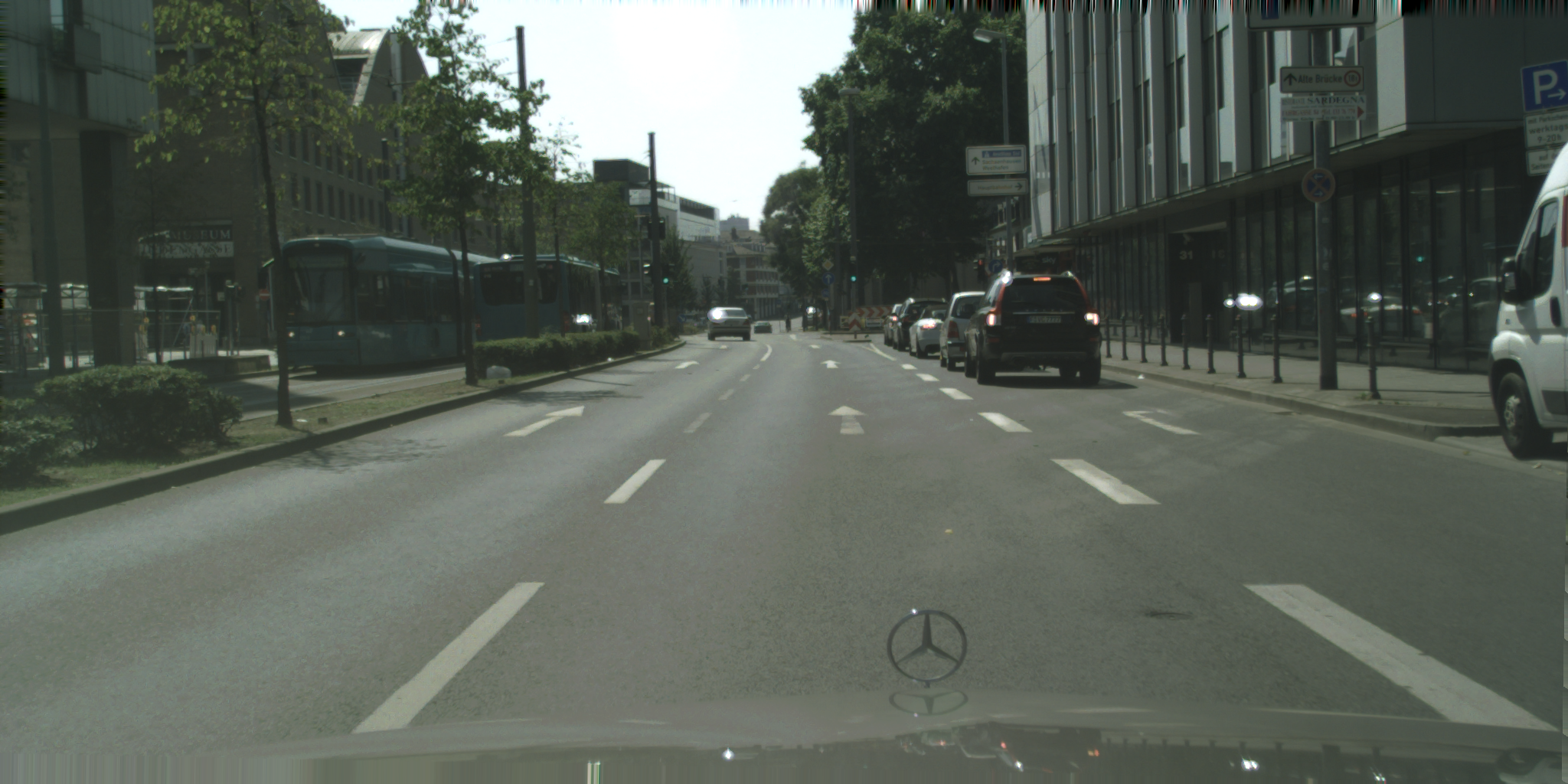}
	\end{subfigure}
	\begin{subfigure}{0.24\textwidth}
		\centering
		\includegraphics[width=4.8cm,height=1.4cm]{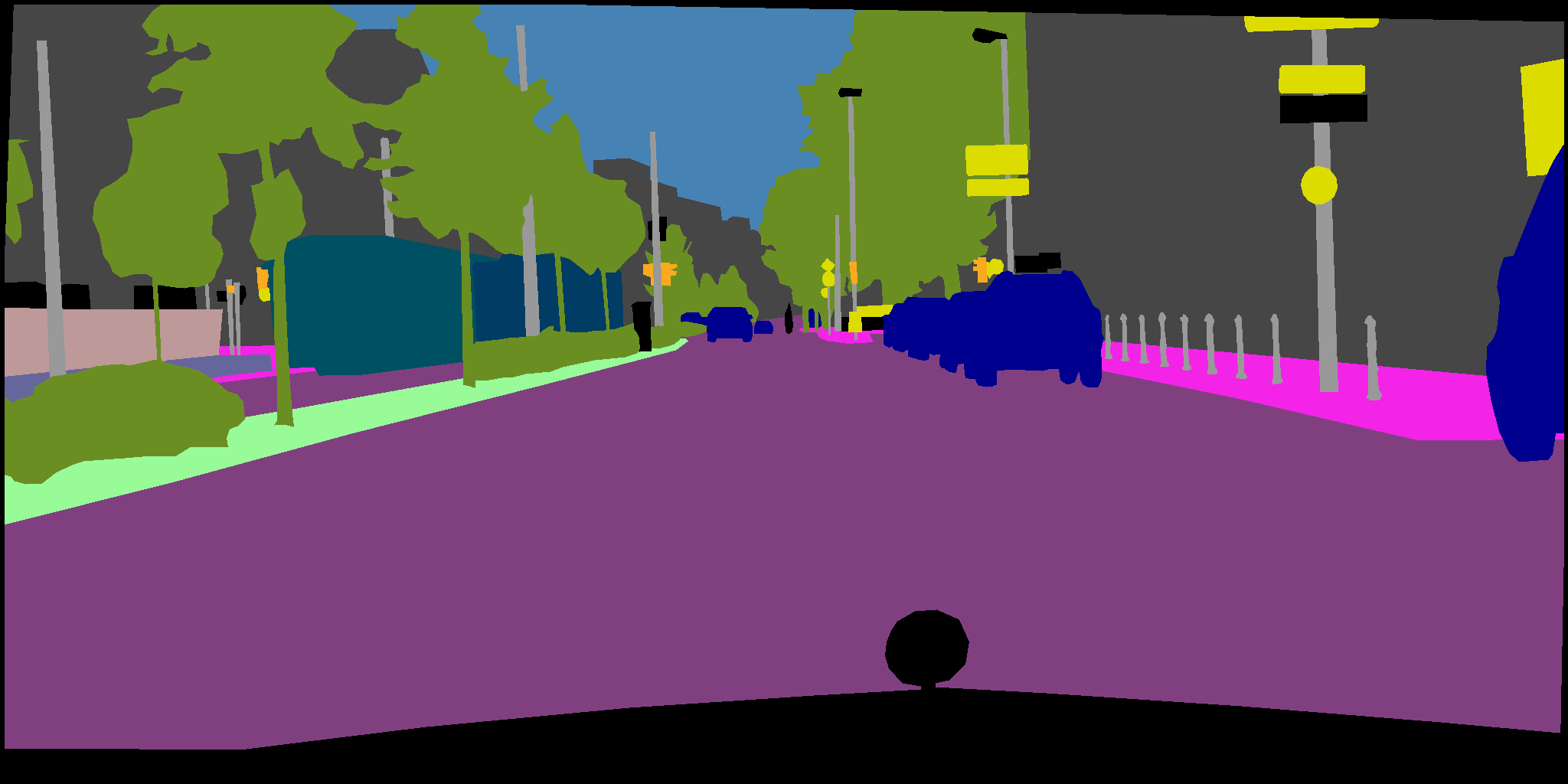}
	\end{subfigure}
	\begin{subfigure}{0.24\textwidth}
		\centering
		\includegraphics[width=4.8cm,height=1.4cm]{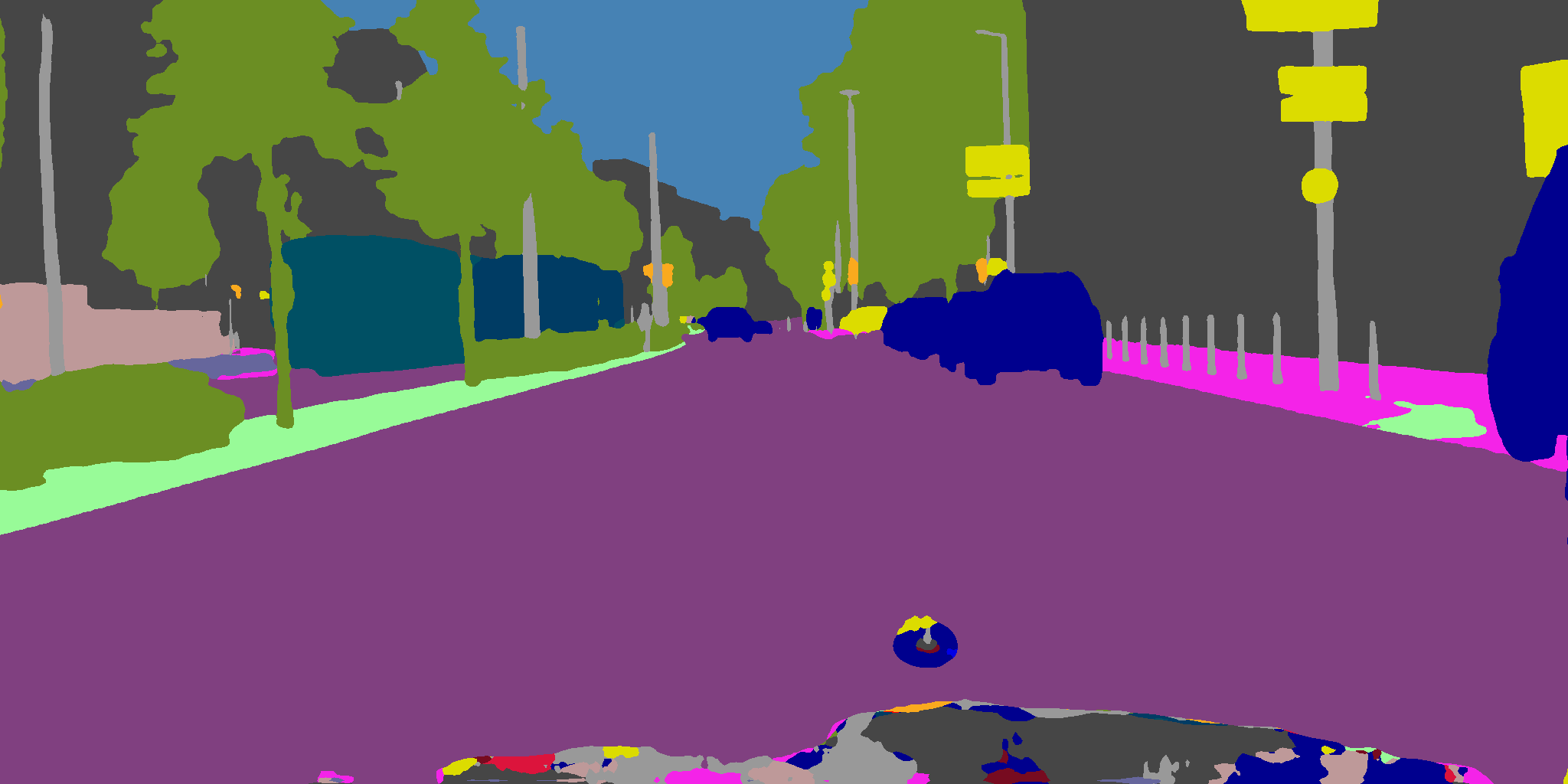}
	\end{subfigure}
	\\
	\centering
	\begin{subfigure}{0.24\textwidth}
		\centering
		\includegraphics[width=4.8cm,height=1.4cm]{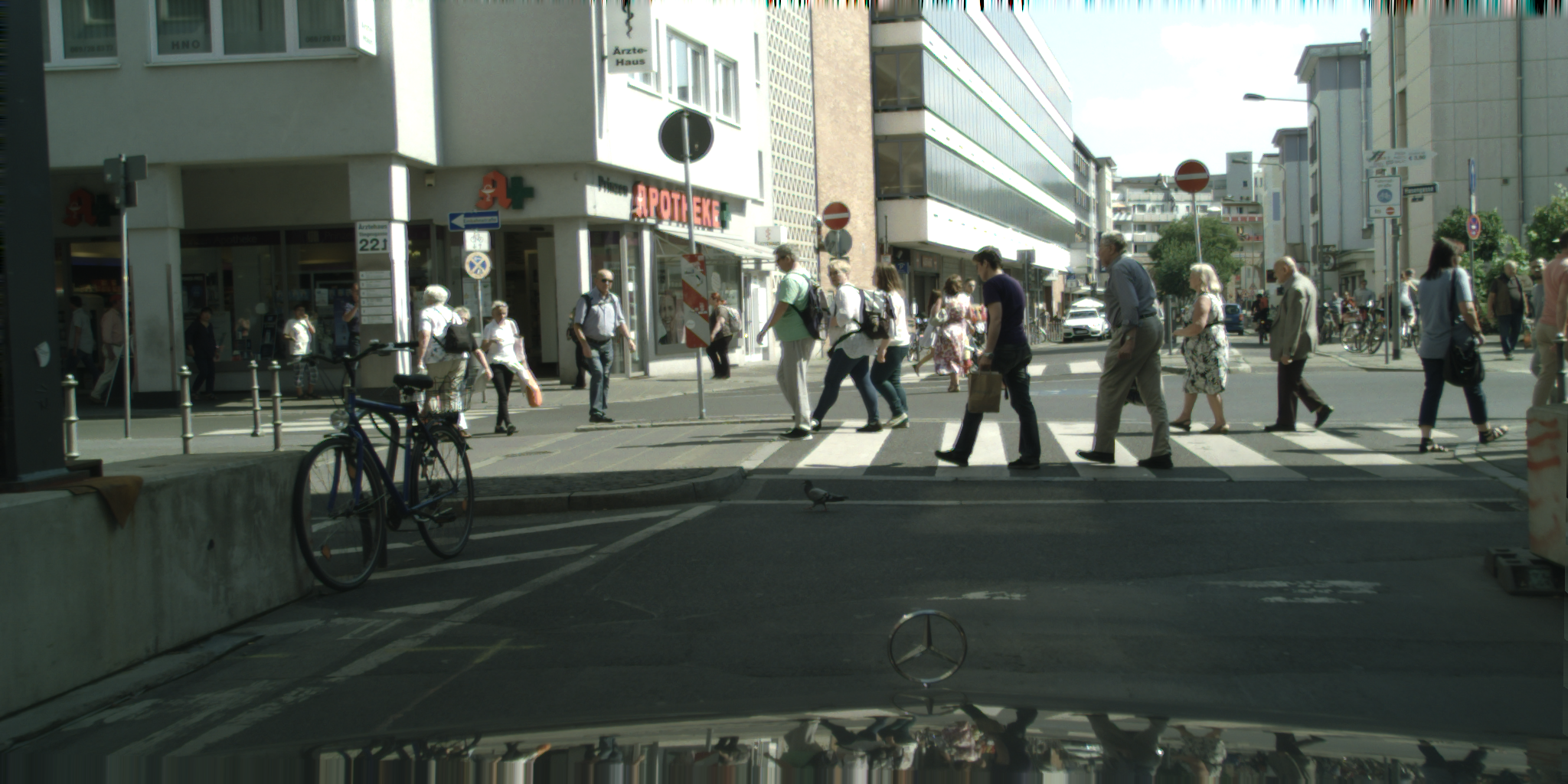}
	\end{subfigure}
	\begin{subfigure}{0.24\textwidth}
		\centering
		\includegraphics[width=4.8cm,height=1.4cm]{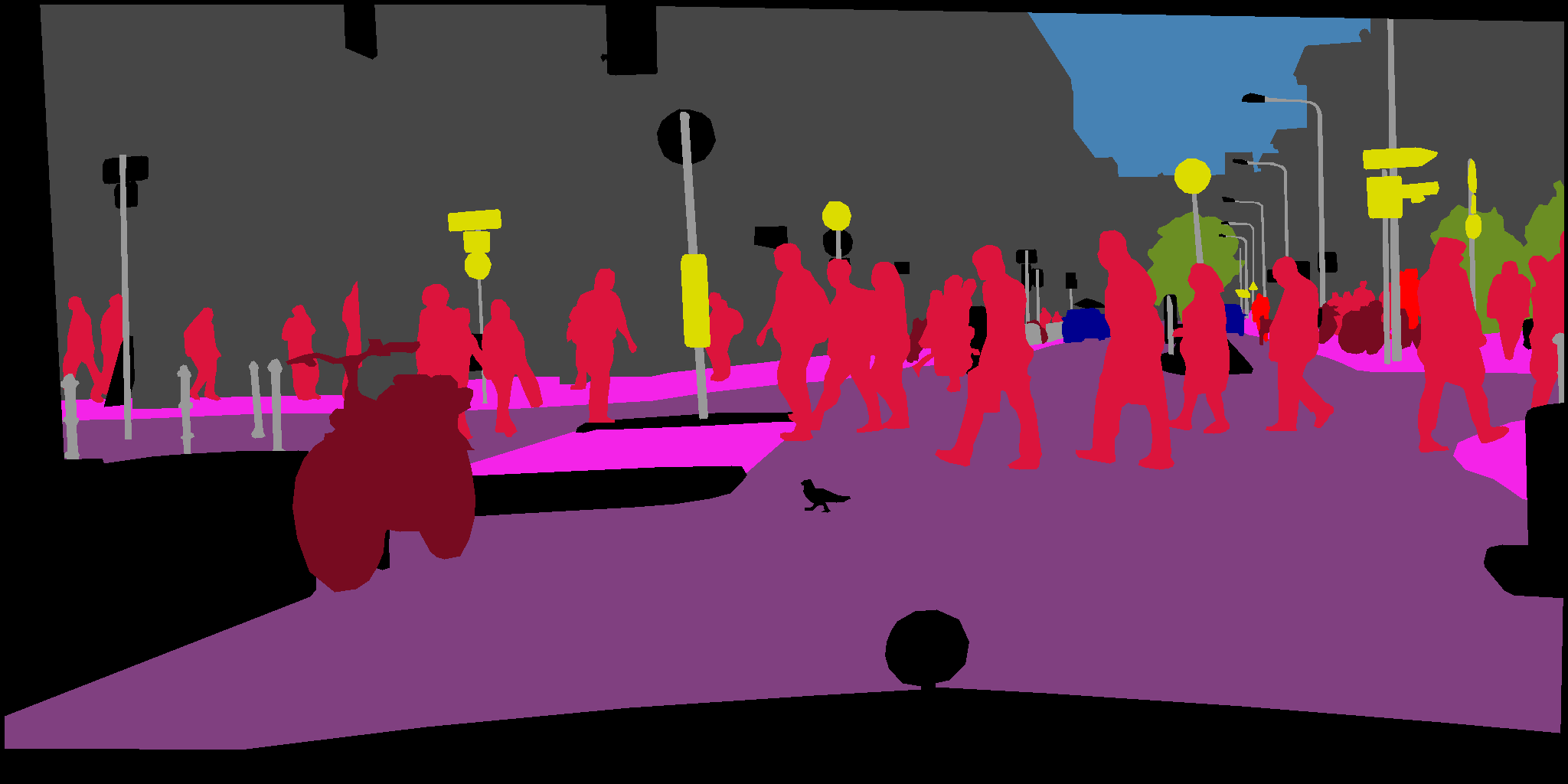}
	\end{subfigure}
	\begin{subfigure}{0.24\textwidth}
		\centering
		\includegraphics[width=4.8cm,height=1.4cm]{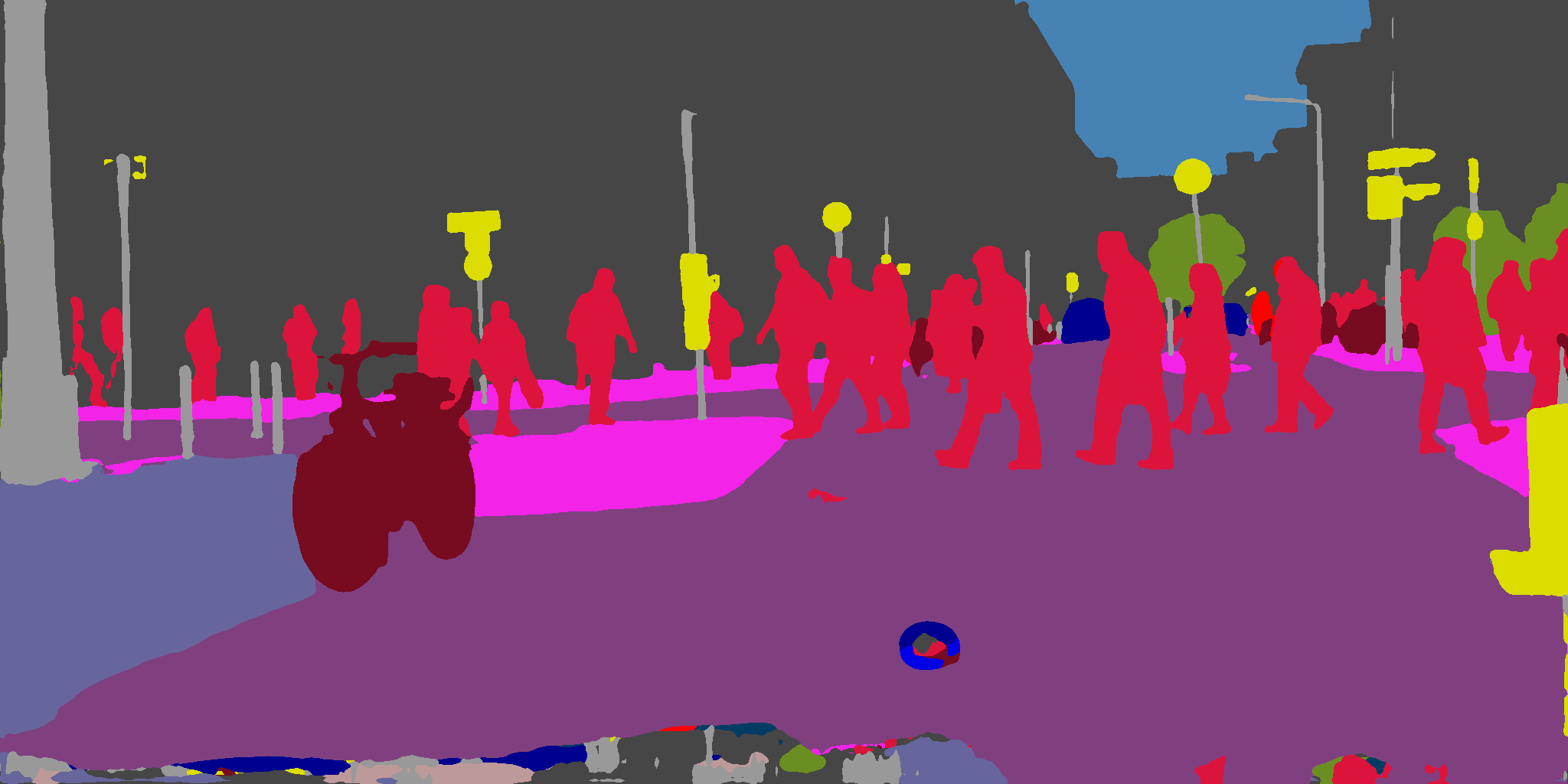}
	\end{subfigure}
	\\
	\centering
	\begin{subfigure}{0.24\textwidth}
		\centering
		\includegraphics[width=4.8cm,height=1.4cm]{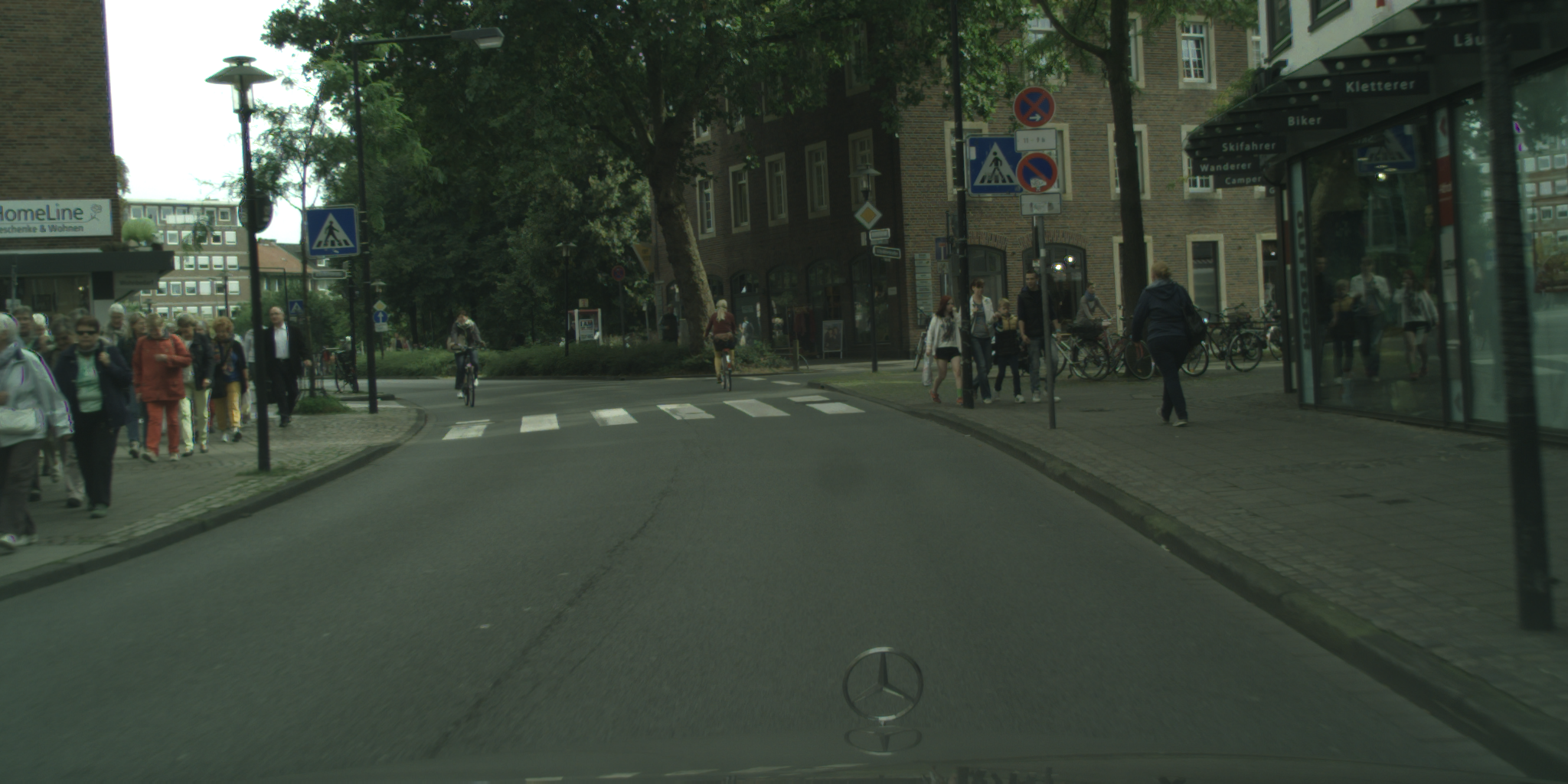}
	\end{subfigure}
	\begin{subfigure}{0.24\textwidth}
		\centering
		\includegraphics[width=4.8cm,height=1.4cm]{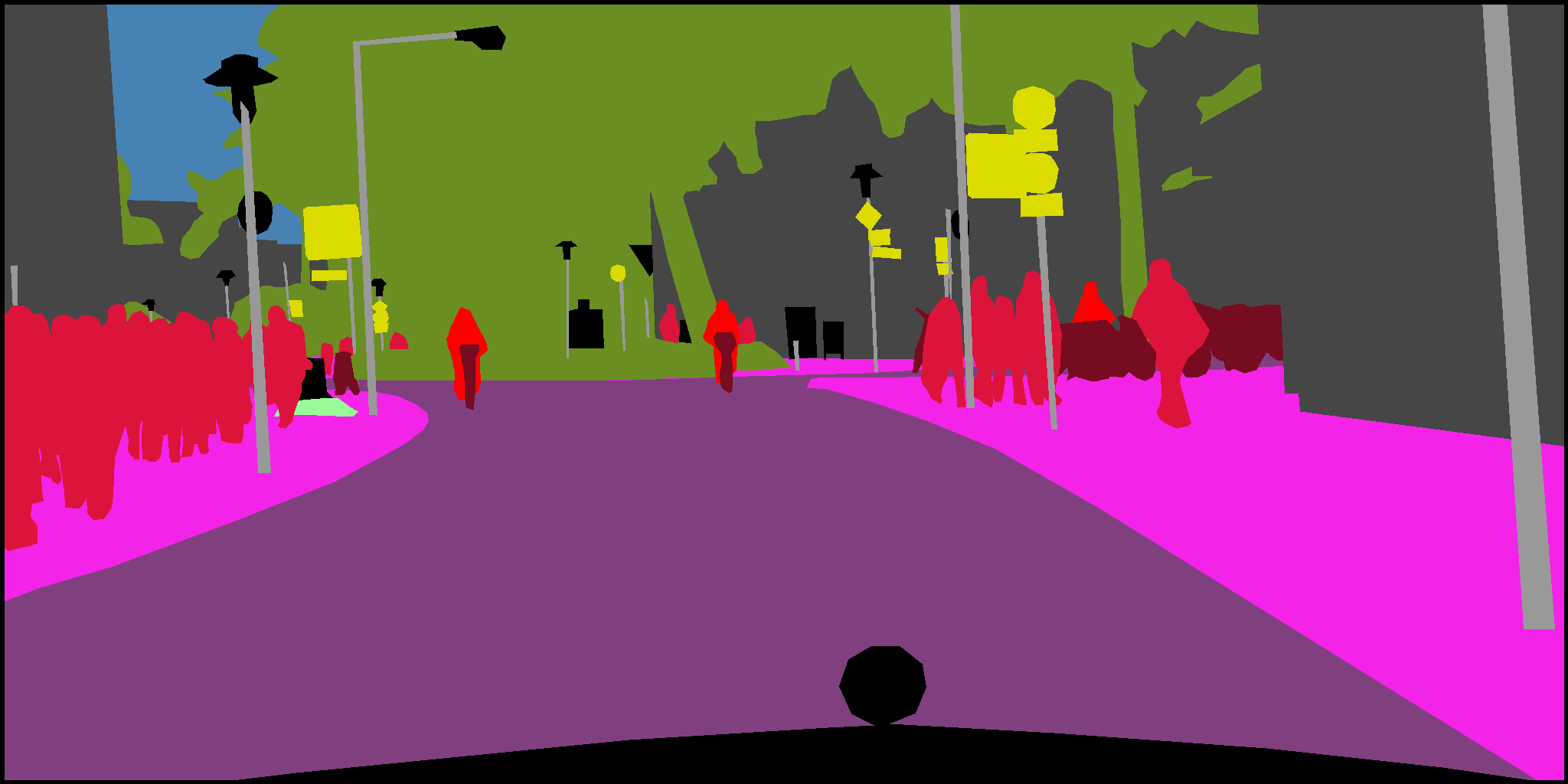}
	\end{subfigure}
	\begin{subfigure}{0.24\textwidth}
		\centering
		\includegraphics[width=4.8cm,height=1.4cm]{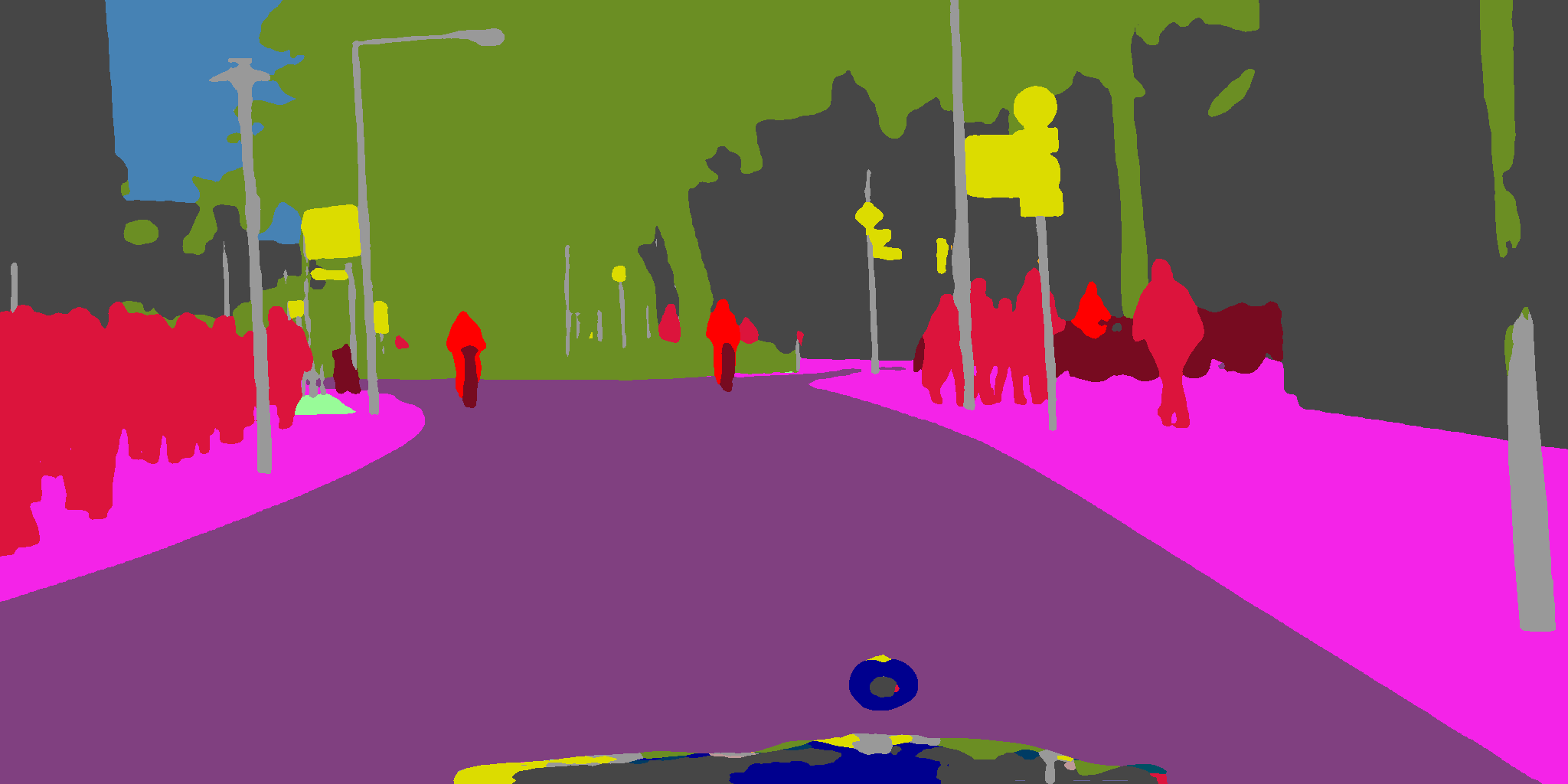}
	\end{subfigure}
	\\
	\centering
	\begin{subfigure}{0.24\textwidth}
		\centering
		\includegraphics[width=4.8cm,height=1.4cm]{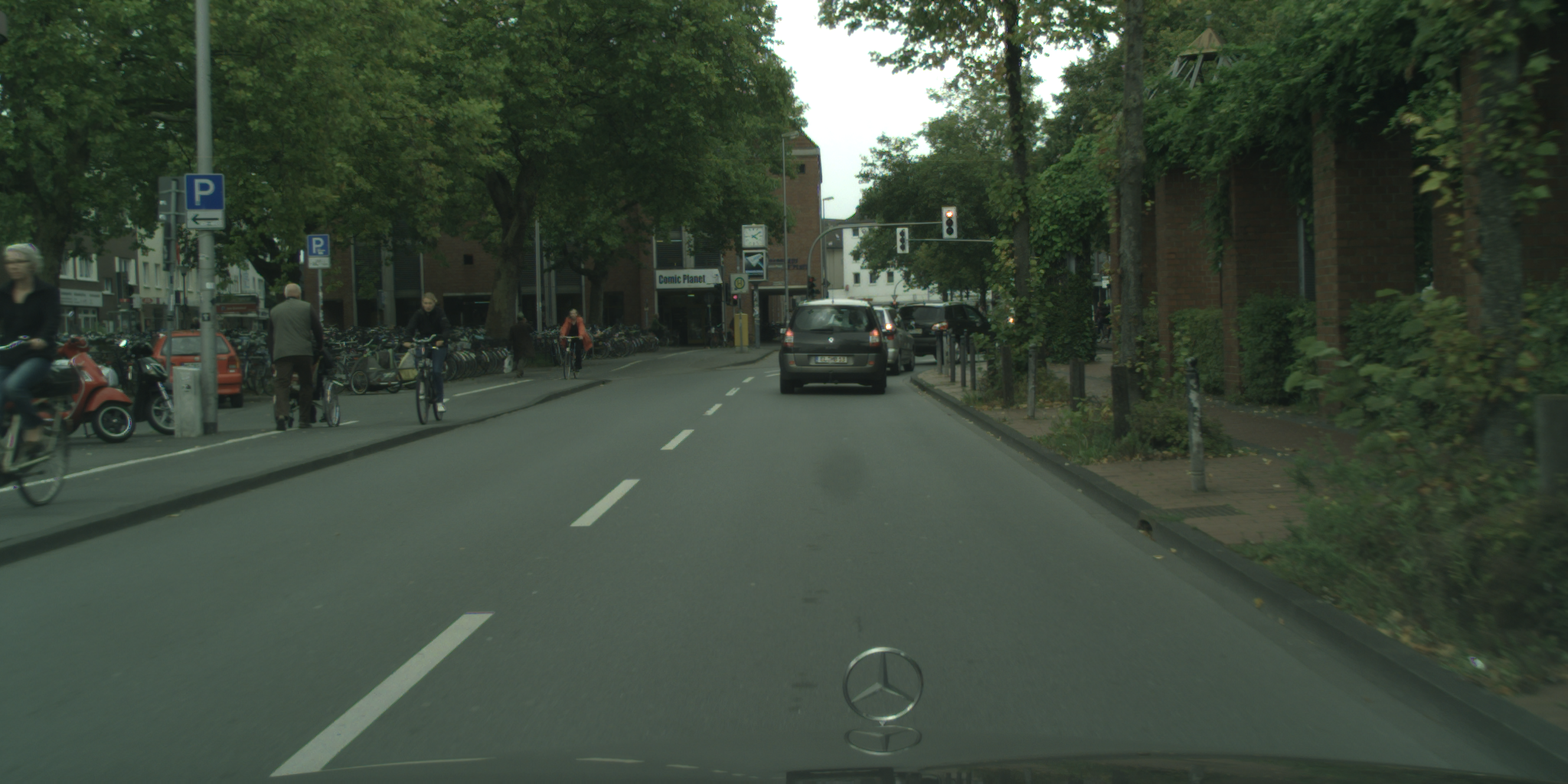}
	\end{subfigure}
	\begin{subfigure}{0.24\textwidth}
		\centering
		\includegraphics[width=4.8cm,height=1.4cm]{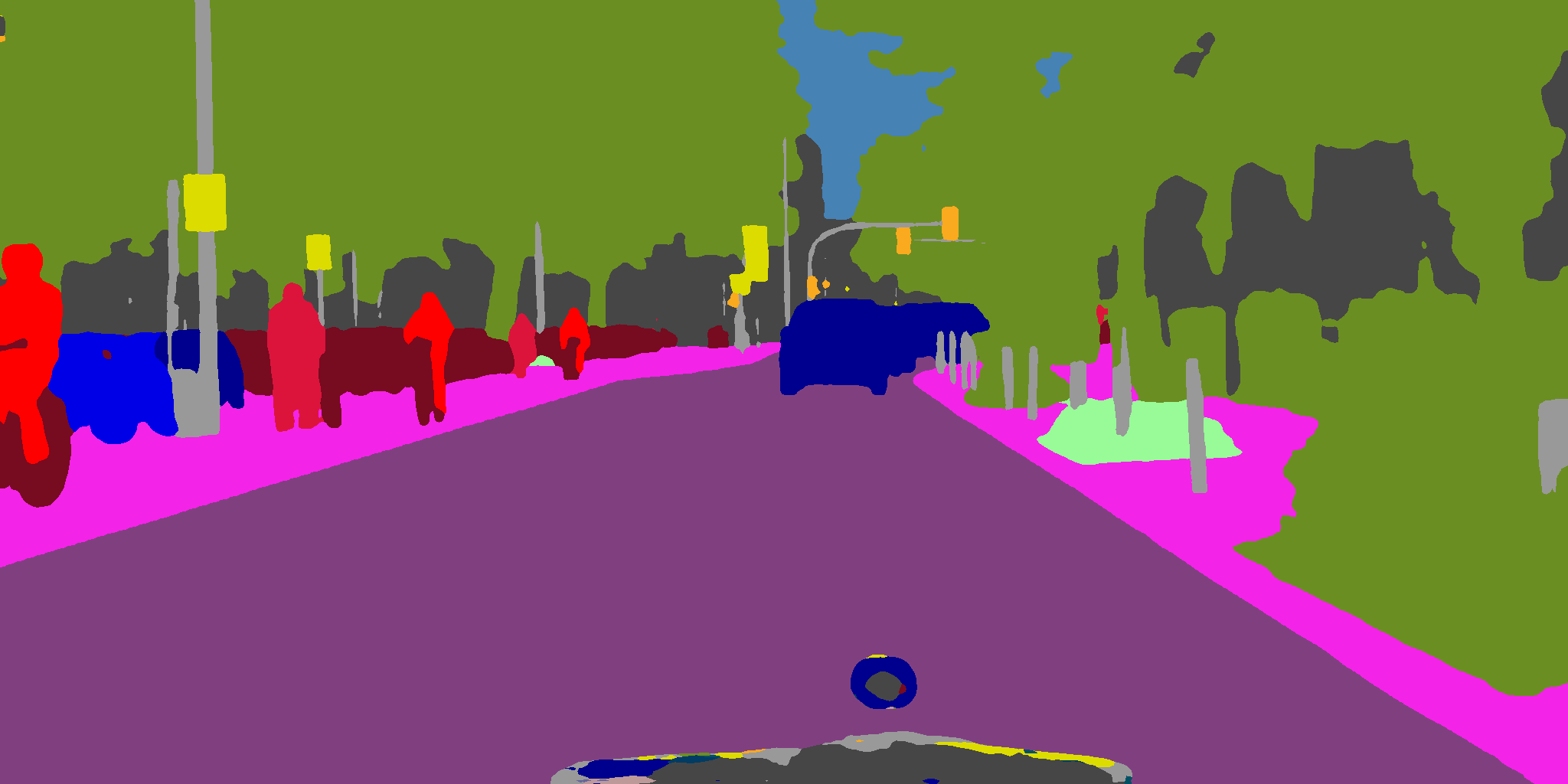}
	\end{subfigure}
	\begin{subfigure}{0.24\textwidth}
		\centering
		\includegraphics[width=4.8cm,height=1.4cm]{images/city05_color.png}
	\end{subfigure}
	\centering
	\\
	\centering
	\begin{subfigure}{0.24\textwidth}
		\centering
		\includegraphics[width=4.8cm,height=1.4cm]{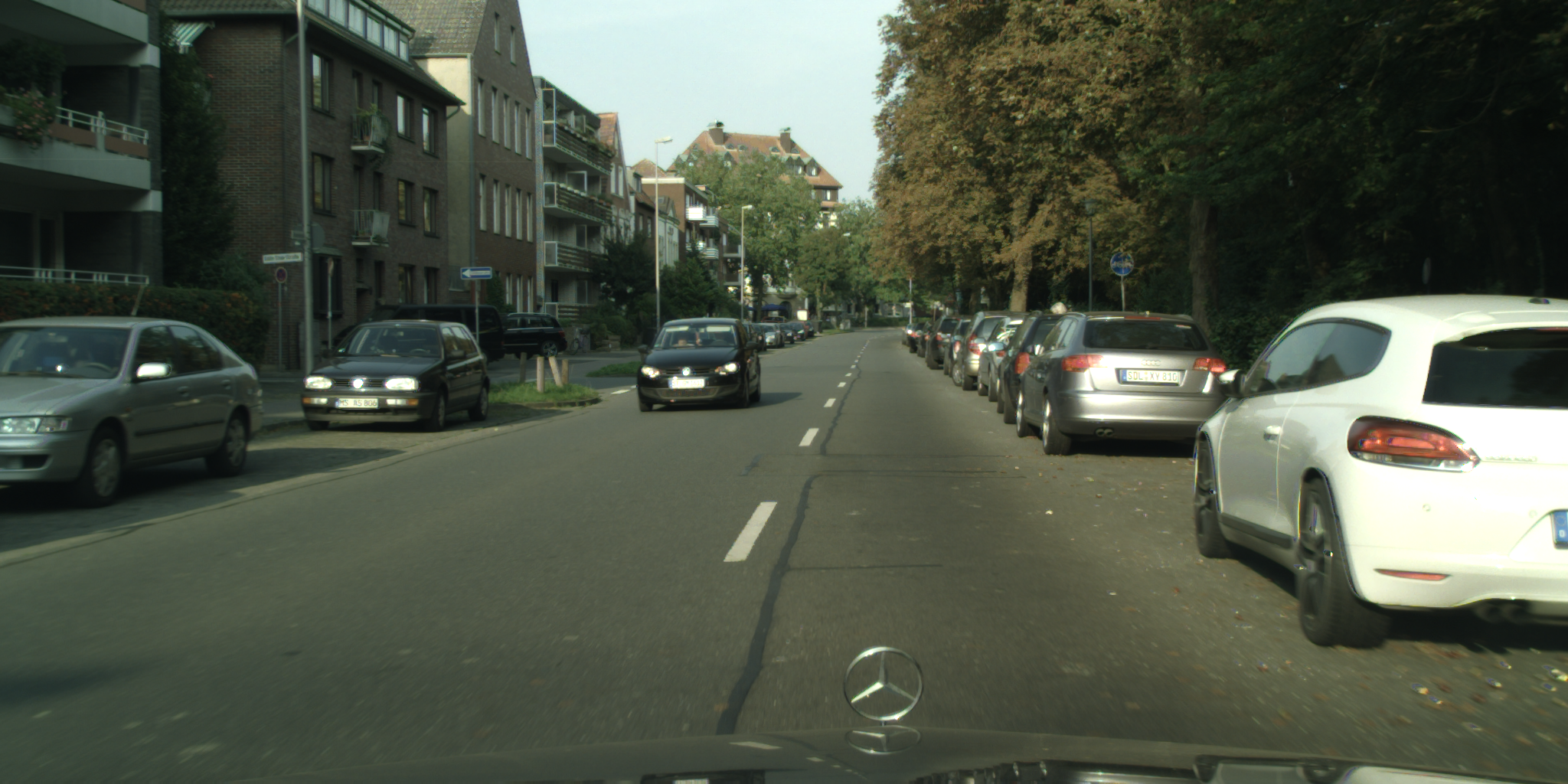}
	\end{subfigure}
	\begin{subfigure}{0.24\textwidth}
		\centering
		\includegraphics[width=4.8cm,height=1.4cm]{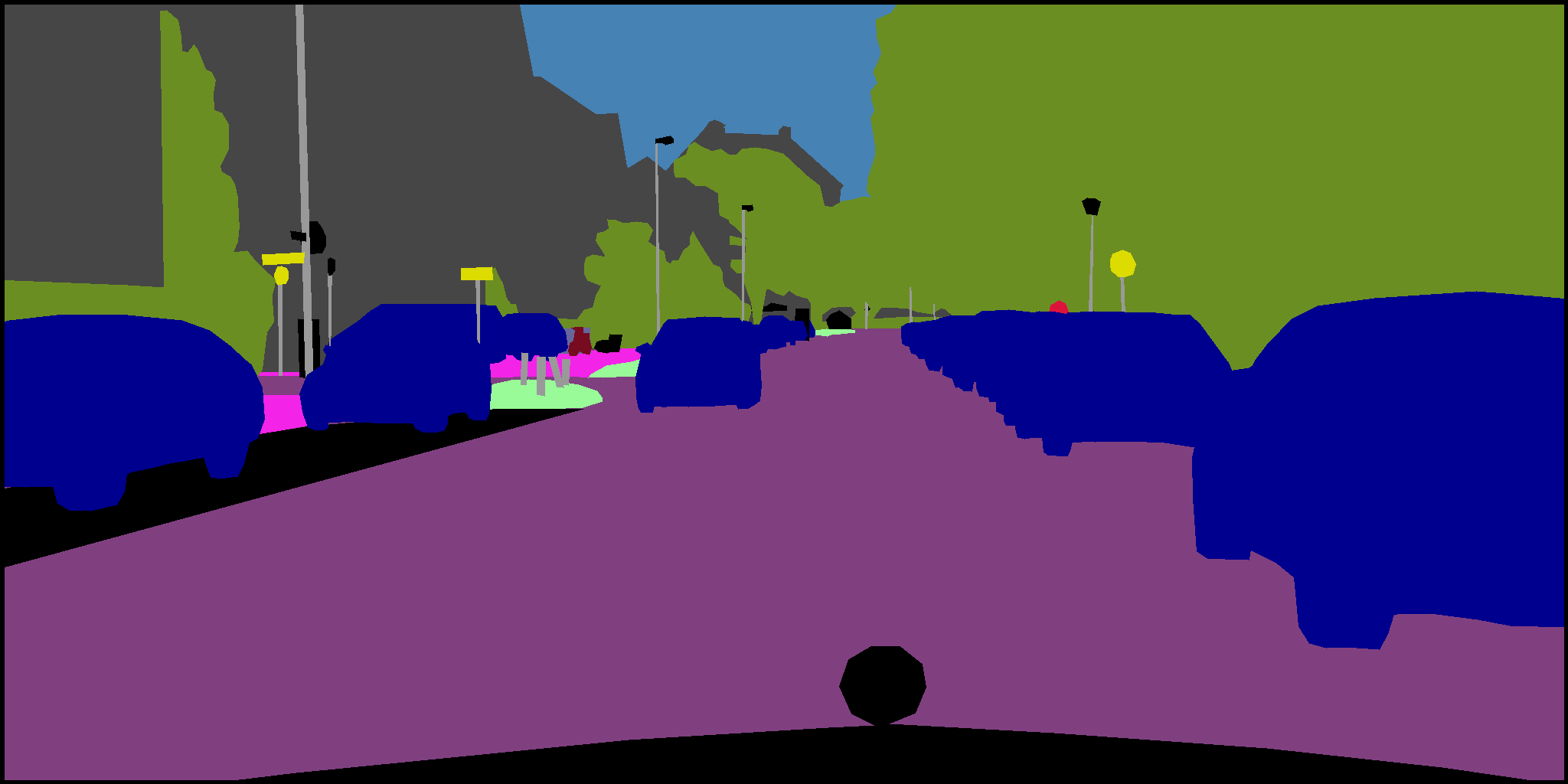}
	\end{subfigure}
	\begin{subfigure}{0.24\textwidth}
		\centering
		\includegraphics[width=4.8cm,height=1.4cm]{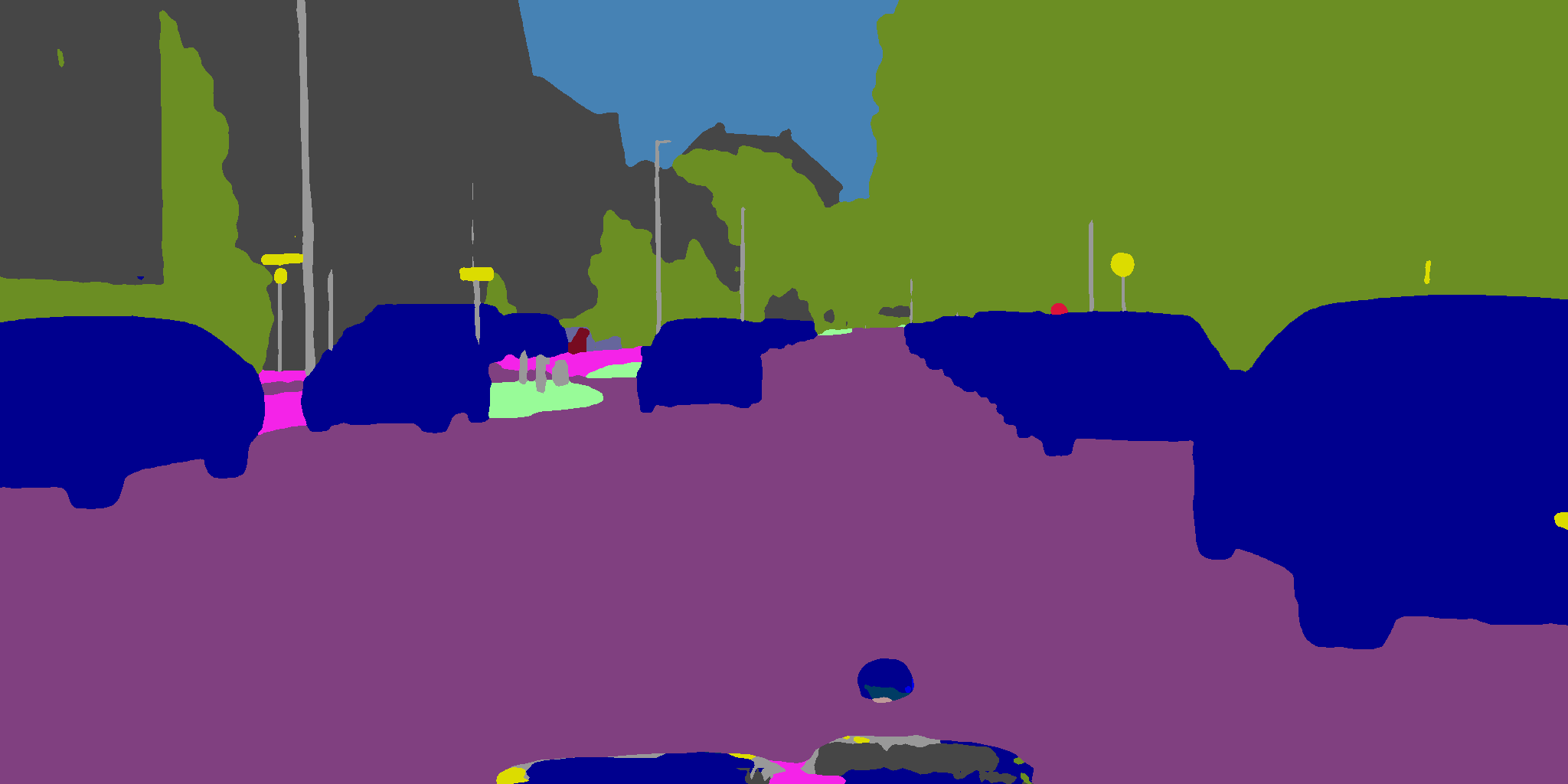}
	\end{subfigure}
	\centering
	\caption{Visual results of our method P2AT-M on Cityscapes dataset. The first row is the original image, the second row is the ground truth, and the last row represents the model performance.}
	\label{cityscapes_fig}
\end{figure*}

\subsection{Results on Camvid Dataset}

In this section, we analyze and discuss the performance of the proposed methods, namely P2AT-S, P2AT-M, and P2AT-L, when compared to other state-of-the-art (SOTA) methods on the Camvid test dataset. The results are presented in Table \ref{camvid_wit_sota}.
First, let's consider the performance of the SOTA methods on the Camvid dataset. Deeplab \cite{chen2017deeplab} and PSPNet \cite{zhao2017pyramid} achieved an accuracy of 61.6\% mIoU and 69.1\% mIoU, respectively. These methods utilized different approaches, with Deeplab employing dilated convolutions and PSPNet pyramid pooling modules. 
DFANet-A and DFANet-B \cite{li2019dfanet}, introduced in 2019, adopted lightweight architectures with 7.8 and 4.8 million parameters, respectively. DFANet-A achieved 64.7\% mIoU with an inference speed of 120 FPS, while DFANet-B achieved 59.3\% mIoU with a faster speed of 160 FPS. These methods demonstrated a higher inference speed but their accuracy is very low, and they have been further improved upon by subsequent approaches.
GAS \cite{lin2020graph} introduced a graph-guided architecture and achieved 72.8\% mIoU with a processing speed of 153.1 FPS. LBN-AA \cite{dong2020real}, on the other hand, utilized an attention aggregation module and obtained 68.0\% mIoU at a speed of 39.3 FPS. 
BiSeNetV2-L, an extended version of BiSeNetV2, further improved the mIoU to 78.5\% by incorporating larger receptive fields. However, the computational speed for BiSeNetV2-L is not provided. STDC1-Seg and STDC2-Seg \cite{fan2021rethinking} utilized the STDC1 and STDC2 architectures, respectively. STDC1-Seg achieved 73\% mIoU and 73.9\%. HyperSeg-S and HyperSeg-L \cite{nirkin2021hyperseg} employed EfficientNet-B1 as the backbone network and achieved mIoU scores of 78.4\% and 79.1\%, respectively. These methods utilized a resolution of 720$\times$960 and demonstrated better performance while maintaining reasonable computational efficiency. HyperSeg-S operated at 38.0 FPS, while HyperSeg-L operated at 16.6 FPS. These methods demonstrated competitive performance but with varying levels of computational efficiency.
Our proposed method P2AT-S, P2AT-M, and P2AT-L utilized ResNet-18, ResNet-34, and ResNet-50 backbones with resolutions of 720$\times$960. We pretrained P2AT on the Cityscapes dataset. P2AT-S* utilized a ResNet-18 backbone and surpasses HyperSeg and BiSeNetV2-L in terms of accuracy with an impressive test accuracy of 80.5\% at a speed of 120 FPS. P2AT-M* with ResNet-34 achieved of 81.0\% at a speed of 103 FPS. P2AT-L*, utilizing the heavy ResNet-50 backbone, achieved an accuracy of 81.1\% mIoU at a speed of 89 FPS. The pretrained versions of P2AT-S, P2AT-M, and P2AT-L outperformed most of the other SOTA methods in terms of mIoU, highlighting their effectiveness and potential for various real-world applications.
These findings validate the effectiveness of the proposed methods in semantic segmentation tasks and their potential for practical applications in various domains. 

\begin{table}
	\caption{\MakeUppercase{The comparison of P2AT various versions on Camvid Dataset.$^{*}$ indicates the models pre-trained on Cityscapes dataset.}}
	\label{camvid_wit_sota}
	\begin{center}
		\begin{adjustbox}{width=0.48\textwidth}
			\small
			\begin{tabu}{lccccc}
				\hline
				\bf{Method}&\bf{backbone}&\bf{Resolution}&\bf{Params}&\bf{Speed (FPS)}&\bf{mIoU\%}\\
				\hline\hline
				Deeplab\cite{chen2017deeplab}&ResNet-101&2017&262.1&4.9&61.6\\
				PSPNet \cite{zhao2017pyramid}&ResNet-101&2017& 250.8&5.4&69.1\\
				\hline
				DFANet-A \cite{li2019dfanet}&XceptionA &2019&7.8&120&64.7\\
				DFANet-B \cite{li2019dfanet}&XceptionB&2019&\textbf{4.8}&160&59.3\\
				\hline
				GAS \cite{lin2020graph}&no&720$\times$960&-&153.1&72.8\\
				LBN-AA \cite{dong2020real}&&360$\times$480&6.2&39.3&68.0\\
				\hline
                BiSeNetV1 \cite{yu2018bisenet}&Xception39&720$\times$960&5.8&175&65.7\\
				BiSeNetV2 \cite{yu2018bisenet}&ResNet-18&720$\times$960&49.0&116.3&68.7\\
				BiSeNetV2\cite{yu2021bisenet}&no&720$\times$960&-&124.5&76.7\\
                BiSeNetV2-L*\cite{yu2021bisenet}&no&720$\times$960&32.7&-&78.5\\
				\hline
				STDC1-Seg \cite{fan2021rethinking}&STDC1&720$\times$960&8.4&\textbf{197.6}&73.0\\
				STDC2-Seg \cite{fan2021rethinking}&STDC2&720$\times$960&12.5&152.2&73.9\\
				\hline 

	            HyperSeg-S\cite{nirkin2021hyperseg}&EfficientNet-B1&720$\times$960&9.9&38.0&78.4\\
				HyperSeg-L \cite{nirkin2021hyperseg}&EfficientNet-B1&720$\times$960&10.2&16.6&79.1\\
                \hline
                P2AT-S* &ResNet-18&720$\times$960&12.6&113.6&80.5\\
				P2AT-M* &ResNet-34&720$\times$960&22.7&86.6&81.0\\
				P2AT-L* &ResNet-50&720$\times$960&37.5&56.1&\textbf{81.1}\\\hline
			\end{tabu}
		\end{adjustbox}
	\end{center}
\end{table}


\begin{table*}
	\caption{\MakeUppercase{Individual category results on Camvid test set in terms of mIoU for 11 classes.
	”-” indicates the corresponding result is not reported by the methods.}}
	\label{tab:individual_camvid_s}
    \begin{adjustbox}{width=0.95\textwidth}
	\vspace{1ex}
	\begin{tabu}{lcccccccccccc}
		\hline
		Method&Building&Tree&Sky&Car&Sign&Road&Ped&Fence&Pole&Sidewalk&Bicyclist&mIoU\\
		\hline
	\hline
	SegNet\cite{badrinarayanan2017segnet}&88.8&\textbf{87.3}&92.4&82.1&20.5&97.2&57.1&49.3&27.5&84.4&30.7&65.2\\
	BiSeNet1\cite{yu2018bisenet}&82.2&74.4&91.9&80.8&42.8&93.3&53.8&49.7&25.4&77.3&50.0&65.6\\
	BiSeNet2\cite{yu2018bisenet}&83.0&75.8&92.0&83.7&46.5&94.6&58.8&53.6&31.9&81.4&54.0&68.7\\
	AGLNet\cite{zhou2020aglnet}&82.6&76.1&91.8&87.0&45.3&95.4&61.5&39.5&39.0&83.1&62.7&69.4\\
	LBN-AA\cite{dong2020real}&83.2&70.5&92.5&81.7&51.6&93.0&55.6&53.2&36.3&82.1&47.9&68.0\\
	BiSeNetV2\slash BiSeNetV2L\cite{yu2021bisenet}&-&-&-&-&-&-&-&-&-&-&-&72.4\slash 73.2\\ \hline
	P2AT-S &\textbf{91.6}&81.4&93.3&95.0&55.8&96.7&77.1&\textbf{75.7}&\textbf{50.4}&90.4&74.7&80.5\\ 
	P2AT-M &91.0&81.4&\textbf{93.4}&\textbf{95.3}&55.9&\textbf{97.6}&77.5&73.8&49.3&\textbf{92.6}&77.6&81.0\\ 
	P2AT-L &91.0&81.2&93.2&\textbf{95.3}&\textbf{57.3}&97.3&\textbf{77.6}&74.0&48.9&92.0&\textbf{78.9}&81.1\\
	\hline
	\end{tabu}
 \end{adjustbox}
\end{table*}  

\begin{figure}
	\centering
	\begin{subfigure}{0.12\textwidth}
		\centering
		\includegraphics[width=2.4cm,height=1.5cm]{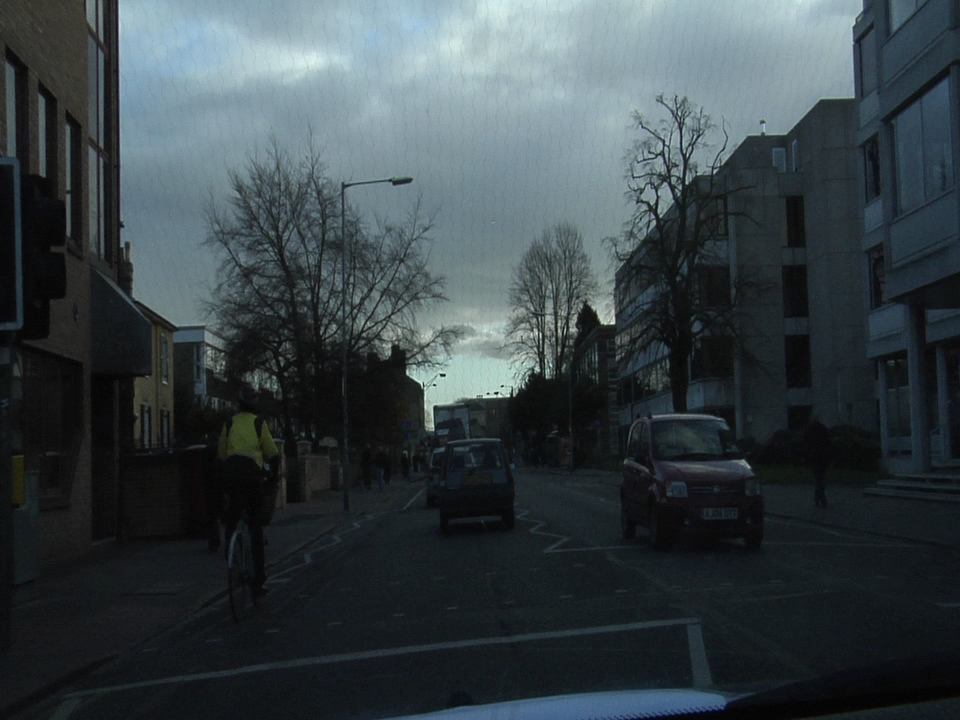}
	\end{subfigure}
	\begin{subfigure}{0.12\textwidth}
		\centering
		\includegraphics[width=2.4cm,height=1.5cm]{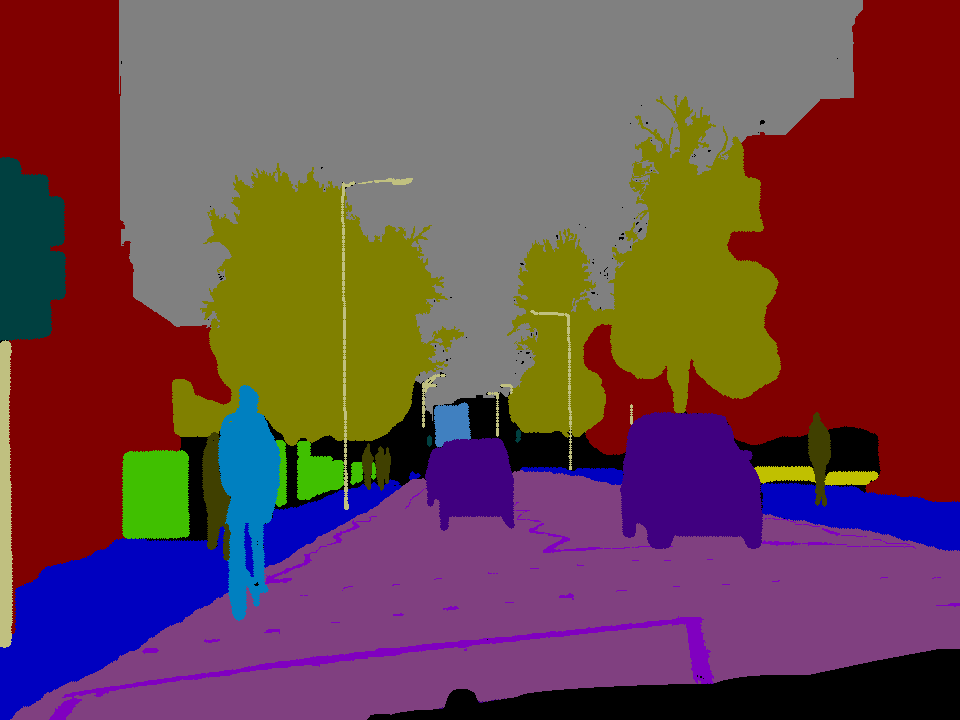}
	\end{subfigure}
	\begin{subfigure}{0.12\textwidth}
		\centering
		\includegraphics[width=2.4cm,height=1.5cm]{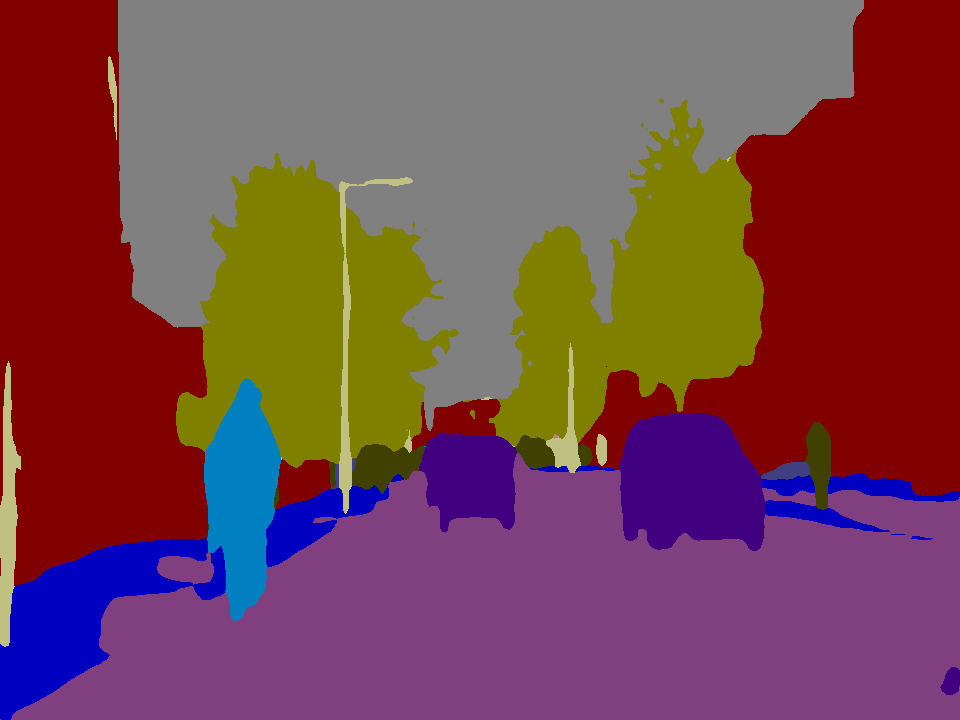}
	\end{subfigure}
	\\
	\begin{subfigure}{0.12\textwidth}
		\centering
		\includegraphics[width=2.4cm,height=1.5cm]{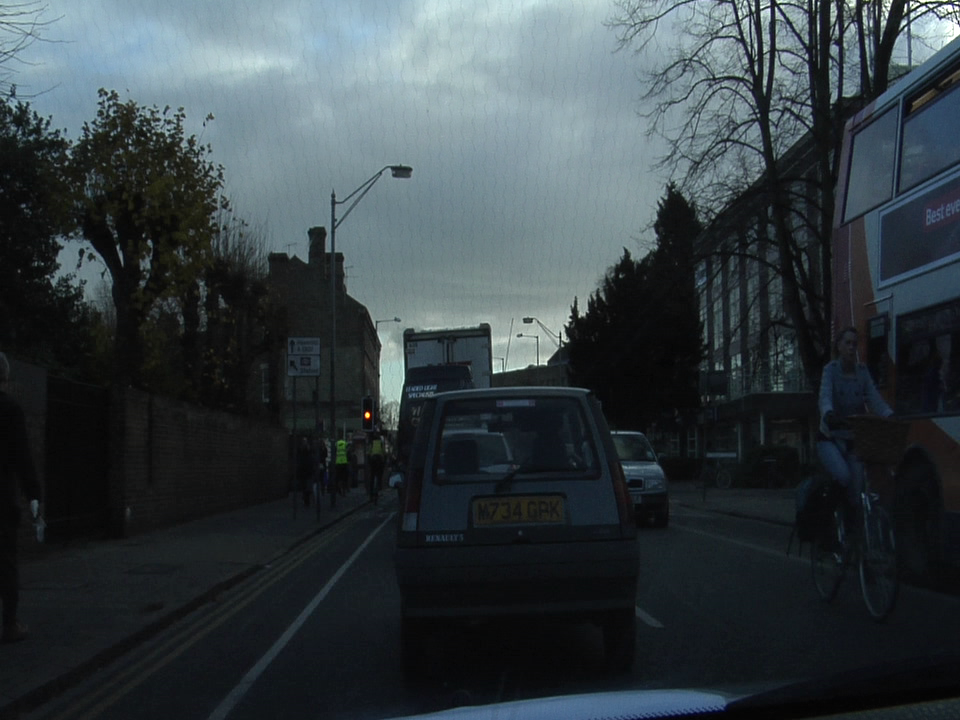}
	\end{subfigure}
	\begin{subfigure}{0.12\textwidth}
		\centering
		\includegraphics[width=2.4cm,height=1.5cm]{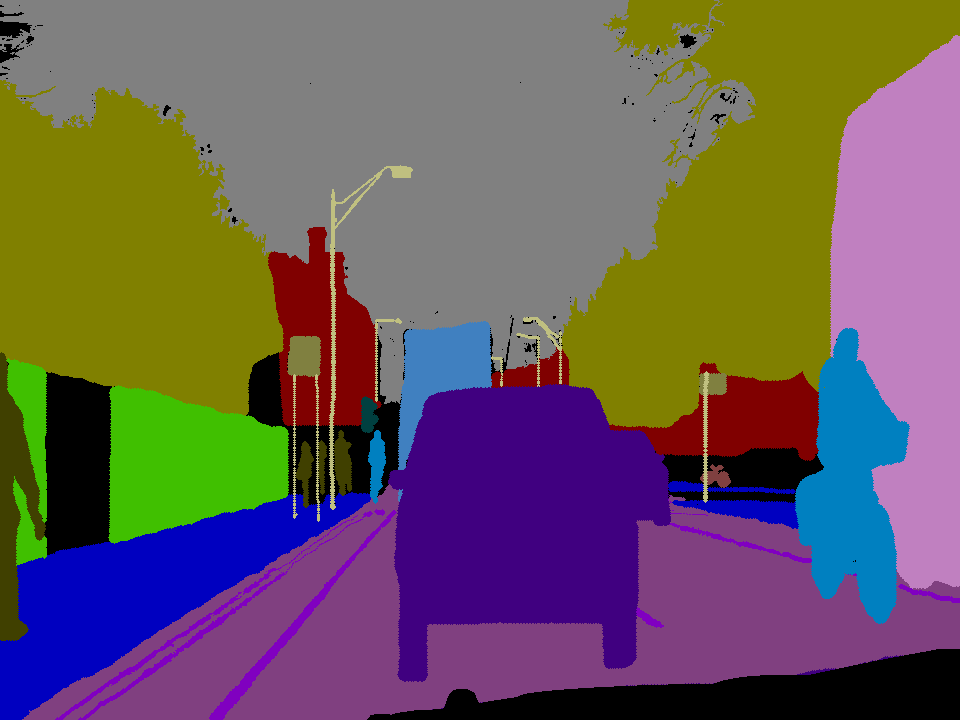}
	\end{subfigure}
	\begin{subfigure}{0.12\textwidth}
		\centering
		\includegraphics[width=2.4cm,height=1.5cm]{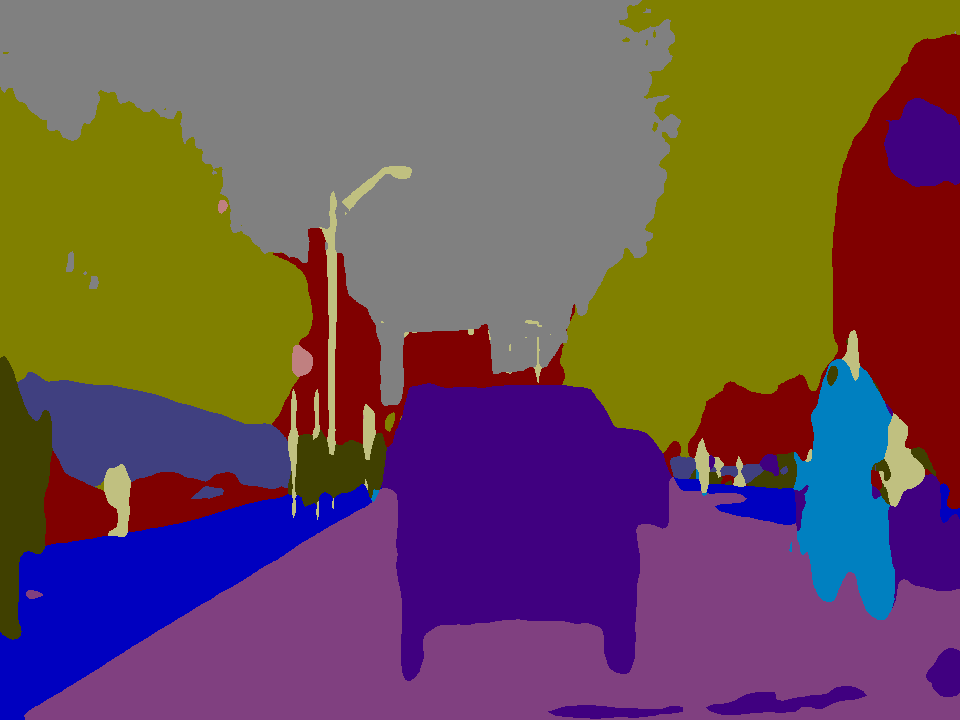}
	\end{subfigure}
	\\
	\centering
	\begin{subfigure}{0.12\textwidth}
		\centering
		\includegraphics[width=2.4cm,height=1.5cm]{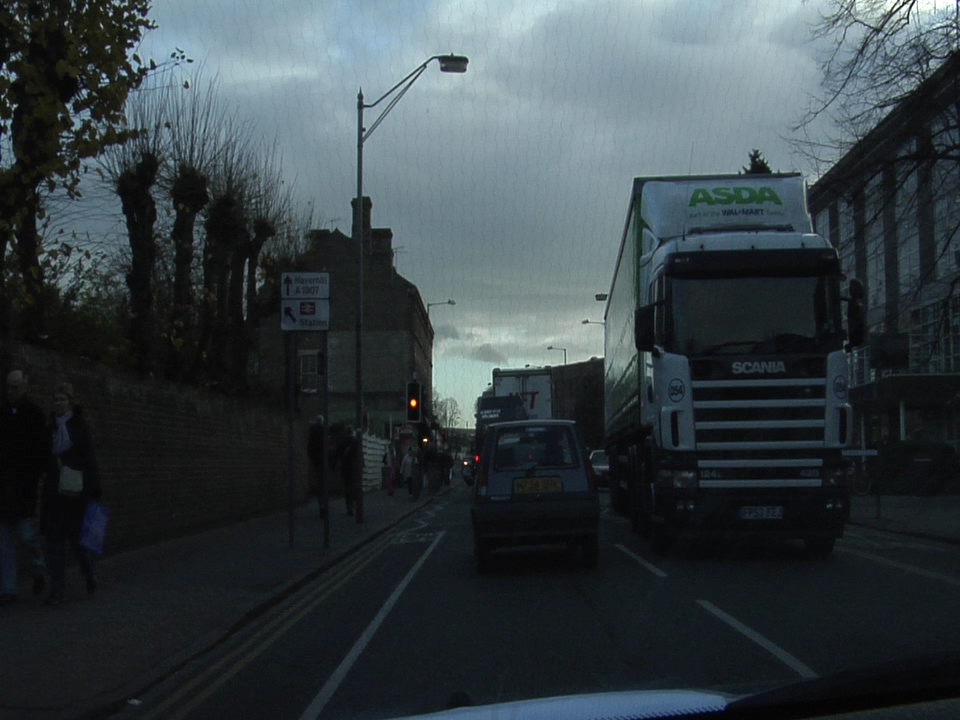}
	\end{subfigure}
	\begin{subfigure}{0.12\textwidth}
		\centering
		\includegraphics[width=2.4cm,height=1.5cm]{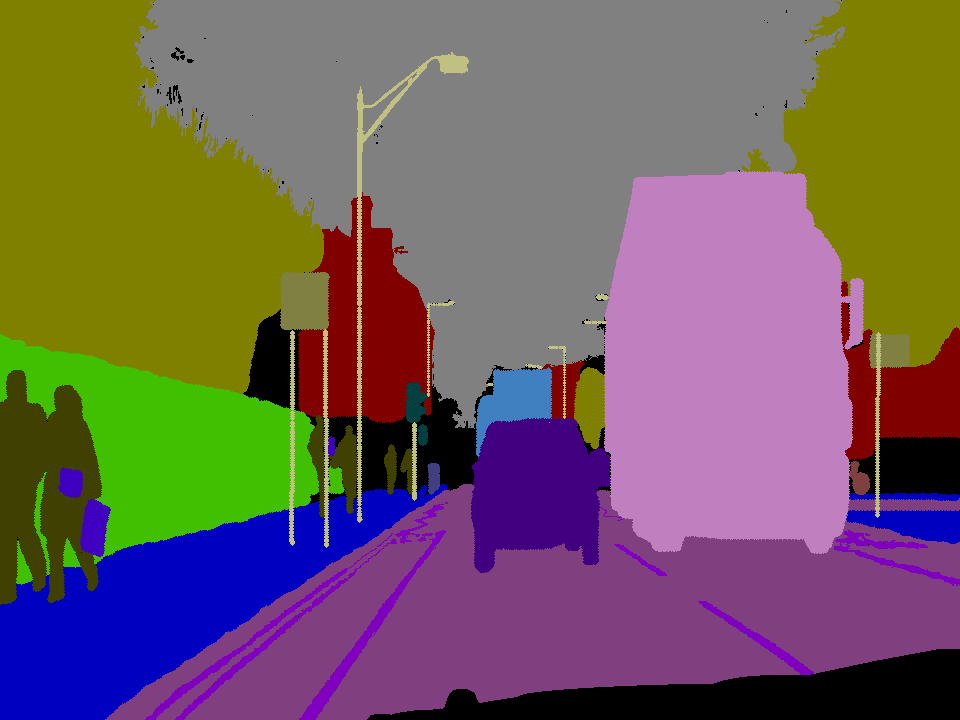}
	\end{subfigure}
	\begin{subfigure}{0.12\textwidth}
		\centering
		\includegraphics[width=2.4cm,height=1.5cm]{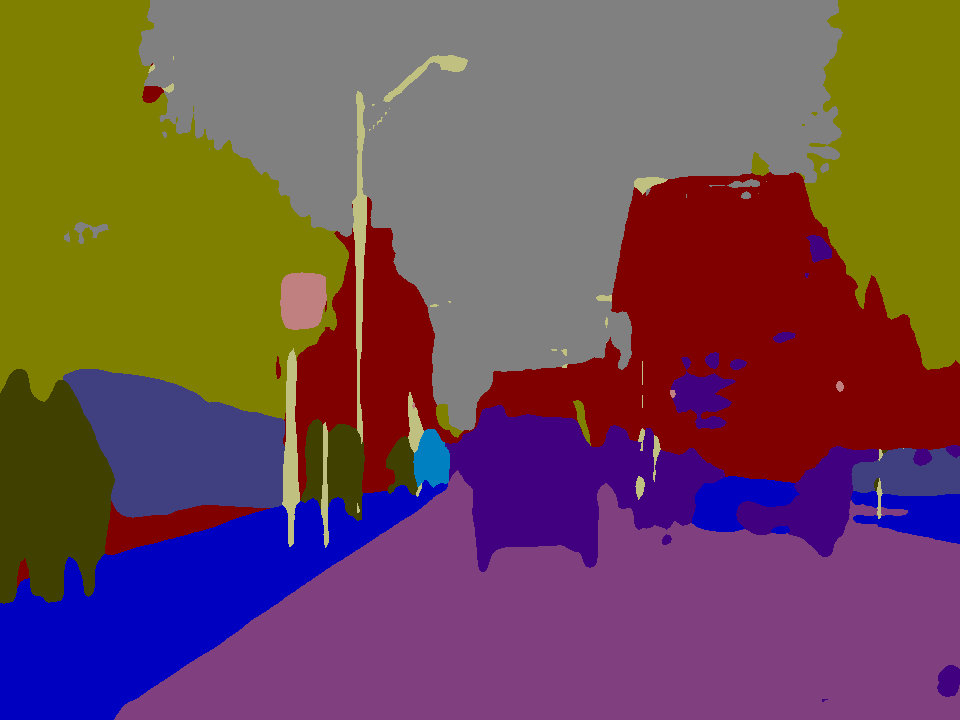}
	\end{subfigure}
	\\
	\centering
	\begin{subfigure}{0.12\textwidth}
		\centering
		\includegraphics[width=2.4cm,height=1.5cm]{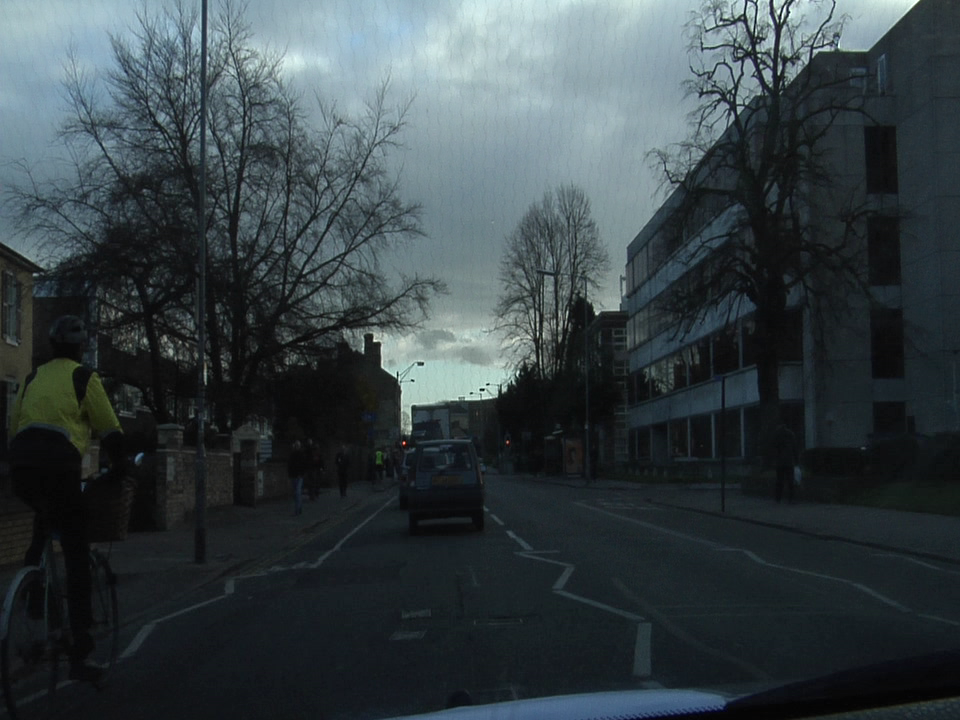}
	\end{subfigure}
	\begin{subfigure}{0.12\textwidth}
		\centering
		\includegraphics[width=2.4cm,height=1.5cm]{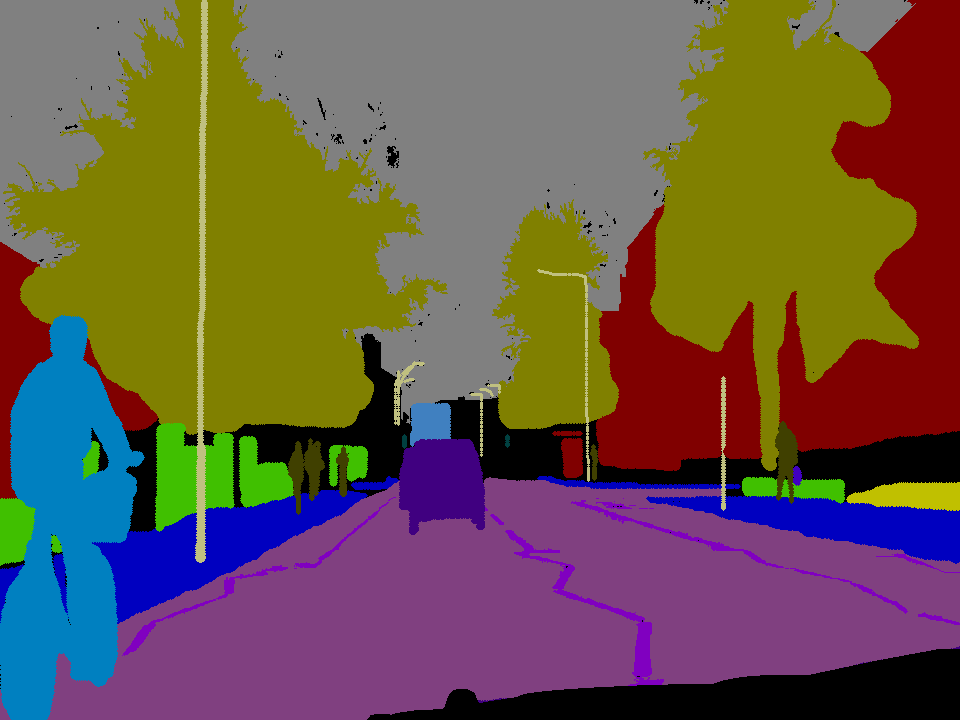}
	\end{subfigure}
	\begin{subfigure}{0.12\textwidth}
		\centering
		\includegraphics[width=2.4cm,height=1.5cm]{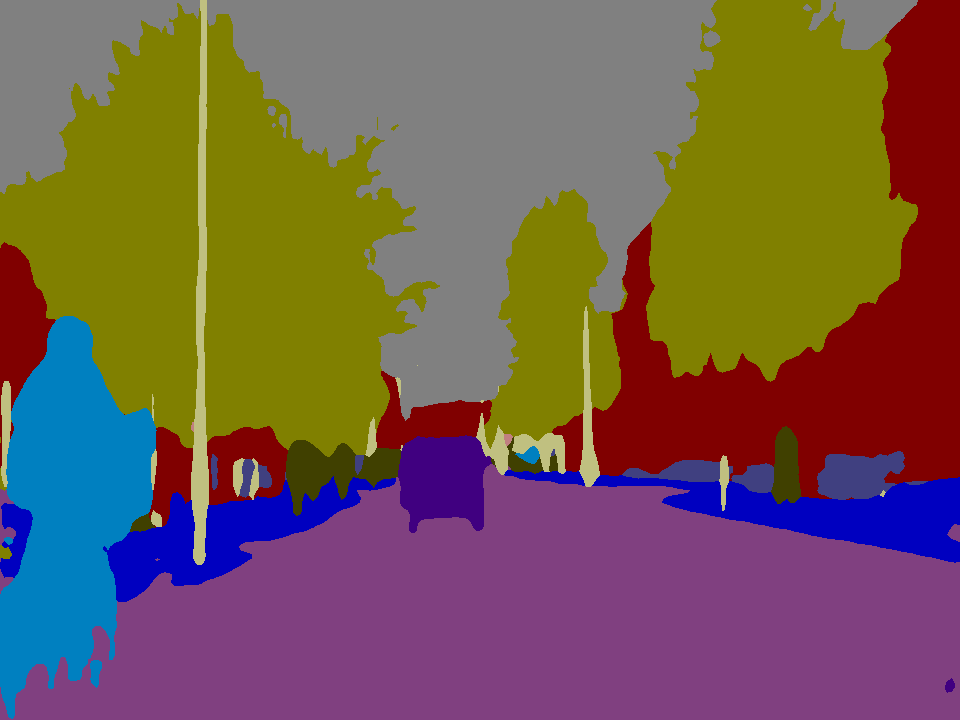}
	\end{subfigure}
	\\
	\centering
	\begin{subfigure}{0.12\textwidth}
		\centering
		\includegraphics[width=2.4cm,height=1.5cm]{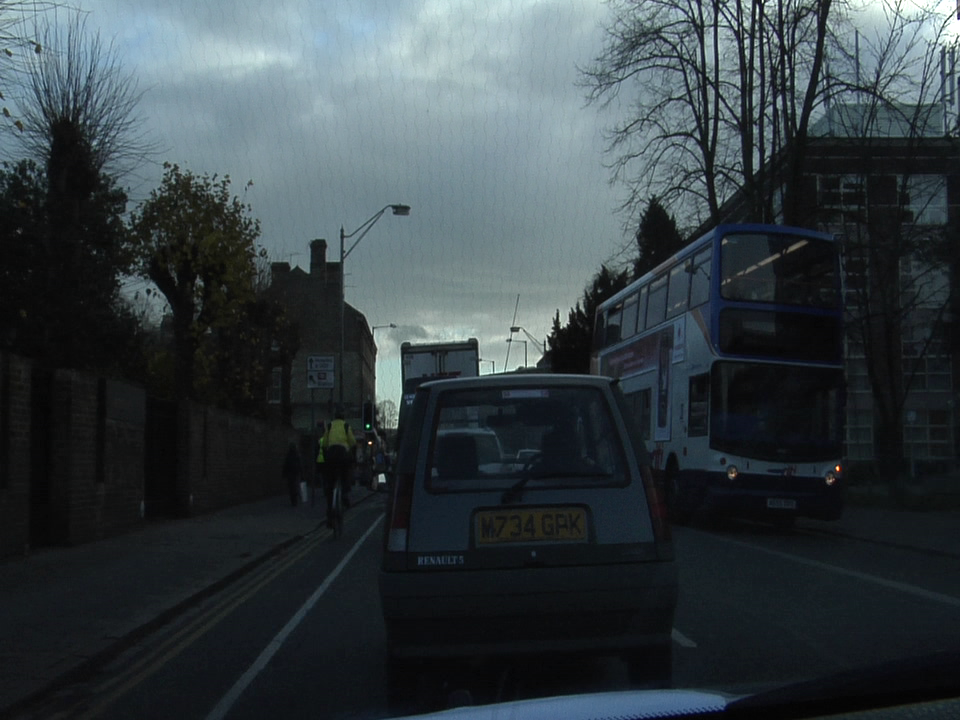}
	\end{subfigure}
	\begin{subfigure}{0.12\textwidth}
		\centering
		\includegraphics[width=2.4cm,height=1.5cm]{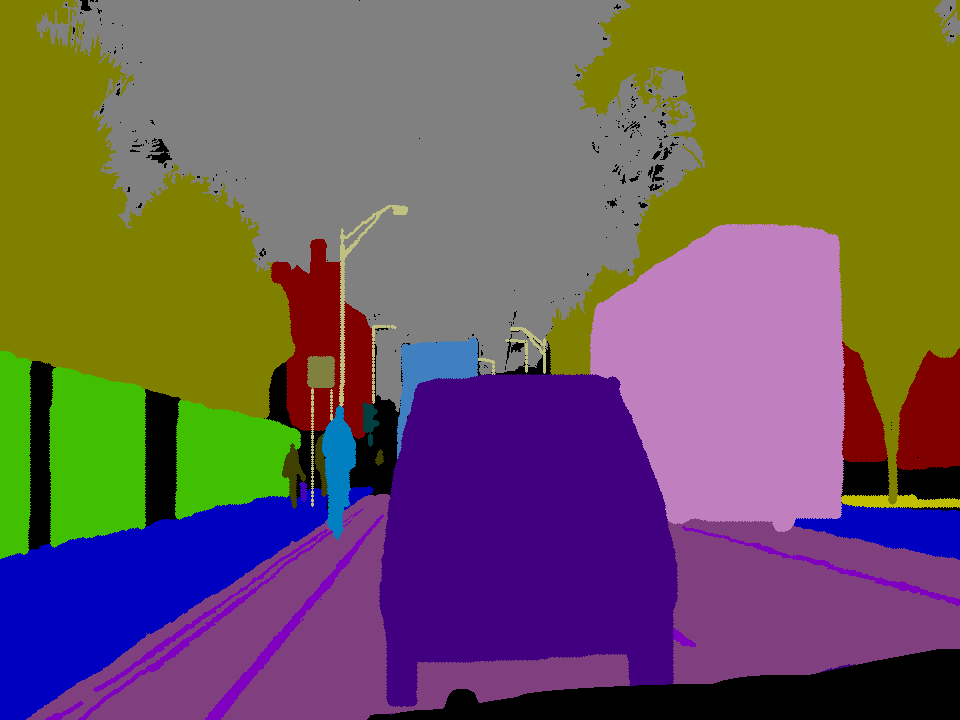}
	\end{subfigure}
	\begin{subfigure}{0.12\textwidth}
		\centering
		\includegraphics[width=2.4cm,height=1.5cm]{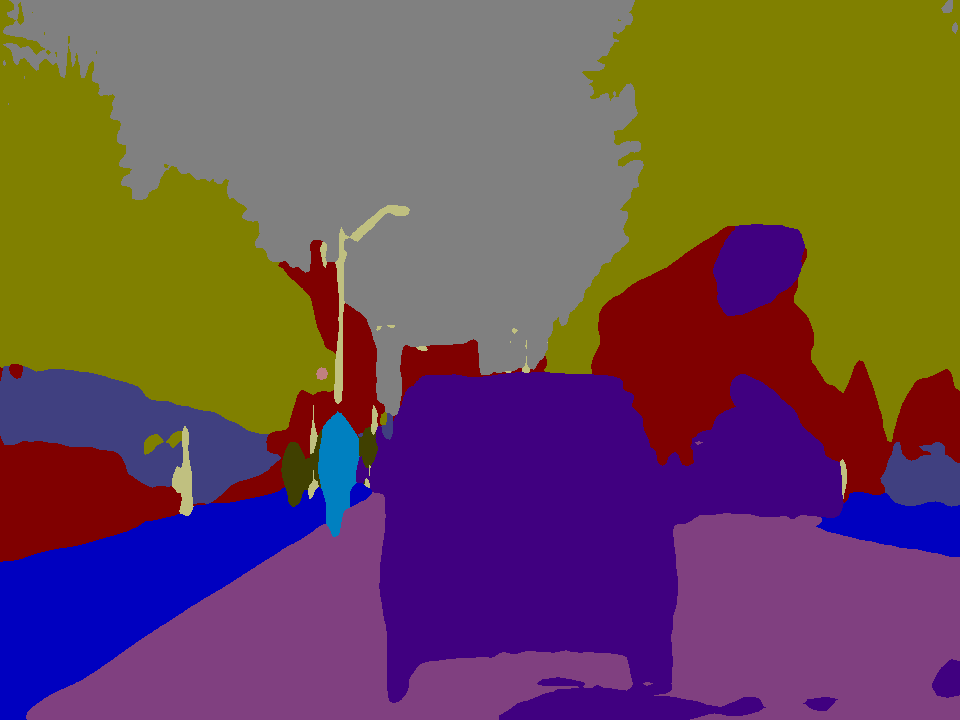}
	\end{subfigure}
	\caption{Visual results of our method P2AT on Camvid test set. The first row is the image, the second row is the prediction, and the last row is the ground truth.}
	\label{Figure:Fig6}
\end{figure}

\begin{figure*}
    \begin{center}
    \begin{subfigure}{0.15\textwidth}
		\includegraphics[width=3.5cm,height=1.5cm]{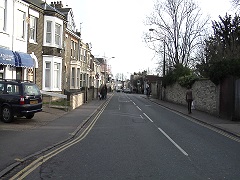}
	\end{subfigure}
	\begin{subfigure}{0.15\textwidth}
		\includegraphics[width=3.5cm,height=1.5cm]{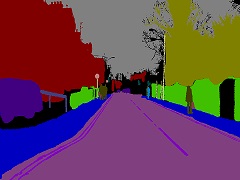}
	\end{subfigure}
	\begin{subfigure}{0.15\textwidth}
		\includegraphics[width=3.5cm,height=1.5cm]{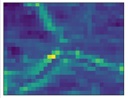}
	\end{subfigure}
	\begin{subfigure}{0.15\textwidth}
		\includegraphics[width=3.5cm,height=1.5cm]{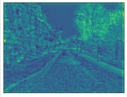}
	\end{subfigure}
        \begin{subfigure}{0.15\textwidth}
		\includegraphics[width=3.5cm,height=1.5cm]{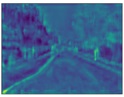}
	\end{subfigure}
        \begin{subfigure}{0.15\textwidth}
		\includegraphics[width=3.4cm,height=1.5cm]{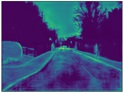}
	\end{subfigure}
         \\
         \begin{subfigure}{0.15\textwidth}
		\includegraphics[width=3.4cm,height=1.5cm]{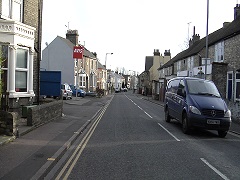}
	\end{subfigure}
	\begin{subfigure}{0.15\textwidth}
		\includegraphics[width=3.4cm,height=1.5cm]{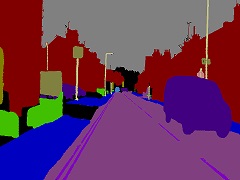}
	\end{subfigure}
	\begin{subfigure}{0.15\textwidth}
		\includegraphics[width=3.4cm,height=1.5cm]{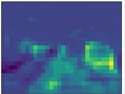}
	\end{subfigure}
	\begin{subfigure}{0.15\textwidth}
		\includegraphics[width=3.4cm,height=1.5cm]{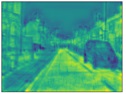}
	\end{subfigure}
        \begin{subfigure}{0.15\textwidth}
		\includegraphics[width=3.4cm,height=1.5cm]{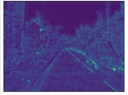}
	\end{subfigure}
          \begin{subfigure}{0.15\textwidth}
		\includegraphics[width=3.4cm,height=1.5cm]{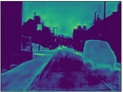}
	\end{subfigure}
        \\
     \begin{subfigure}{0.15\textwidth}
		\includegraphics[width=3.4cm,height=1.5cm]{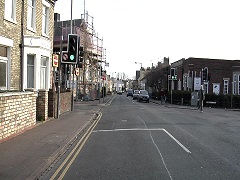}
	\end{subfigure}
	\begin{subfigure}{0.15\textwidth}
		\includegraphics[width=3.4cm,height=1.5cm]{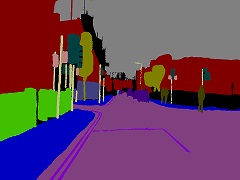}
	\end{subfigure}
	\begin{subfigure}{0.15\textwidth}
		\includegraphics[width=3.4cm,height=1.5cm]{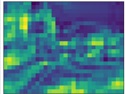}
	\end{subfigure}
	\begin{subfigure}{0.15\textwidth}
		\includegraphics[width=3.4cm,height=1.5cm]{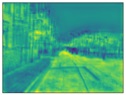}
	\end{subfigure}
        \begin{subfigure}{0.15\textwidth}
		\includegraphics[width=3.4cm,height=1.5cm]{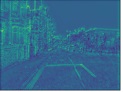}
	\end{subfigure}
         \begin{subfigure}{0.15\textwidth}
		\includegraphics[width=3.4cm,height=1.5cm]{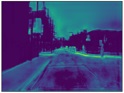}
	\end{subfigure}
         \\
     \begin{subfigure}{0.15\textwidth}
		\includegraphics[width=3.4cm,height=1.5cm]{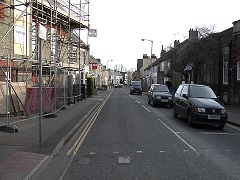}
	\end{subfigure}
	\begin{subfigure}{0.15\textwidth}
		\includegraphics[width=3.4cm,height=1.5cm]{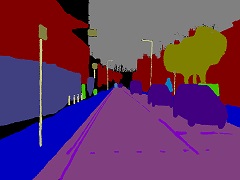}
	\end{subfigure}
	\begin{subfigure}{0.15\textwidth}
		\includegraphics[width=3.4cm,height=1.5cm]{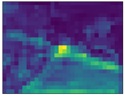}
	\end{subfigure}
	\begin{subfigure}{0.15\textwidth}
		\includegraphics[width=3.4cm,height=1.5cm]{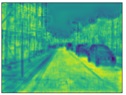}
	\end{subfigure}
        \begin{subfigure}{0.15\textwidth}
		\includegraphics[width=3.4cm,height=1.5cm]{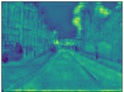}
	\end{subfigure}
        \begin{subfigure}{0.15\textwidth}
		\includegraphics[width=3.4cm,height=1.5cm]{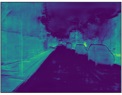}
	\end{subfigure}
	\caption{The feature maps visualization of our proposed architecture on the Camvid validation dataset. From left to right: (a) original input images; (b) The ground truth; (c) The Scale-aware Semantic Aggregation ; (d) the bidirectional fusion (stage4); (e) the global context enhancer (stage 4); (f) the feature refinement before the final segmentation.}
	\label{cityscapes_fig}
 \end{center}
\end{figure*}

\section{Conclusion}
In this work, we have presented P2AT, a real-time semantic segmentation network. The scale-aware context aggregator of P2At is capable of extracting rich contextual information, which is established on our proposed pyramid pooling axial attention. Additionally, we have designed the Bidirectional Fusion (BiF) module that efficiently integrates semantic information at different levels and a global context enhancer module to address limitations in concatenating different semantic levels. Notably, P2AT-S surpasses existing models on the Camvid dataset, achieving an accuracy of 81.0\% mIoU. Moreover, our experiments on Cityscapes and PASCAL VOC 2012 demonstrate the efficiency and effectiveness of the proposed architecture, with P2AT-S and P2AT-L achieving an accuracy of 77.8\%  and 78.0\% on Cityscapes, respectively.
In summary, our contributions encompass the novel P2AT framework, which addresses the challenge of achieving accurate scene understanding in real-time tasks, such as autonomous driving, while considering computational efficiency.

\bibliographystyle{model1-num-names}

\bibliography{cas-refs.bib}

\end{document}